\documentclass{article}

\usepackage{microtype}
\usepackage{graphicx}
\usepackage{caption}
\usepackage{subcaption}
\usepackage{multirow}
\usepackage{wrapfig}
\usepackage{placeins}
\usepackage{booktabs} 
\usepackage{pifont}

\usepackage{hyperref}

\usepackage[accepted]{icml2024}

\usepackage{amsmath}
\usepackage{amssymb}
\usepackage{mathtools}
\usepackage{amsthm}
\usepackage{stfloats}

\newcommand{\Fone}{$F_1$}

\newcommand{\relu}{\mathrm{ReLU}}
\newcommand{\precision}{Pr}
\newcommand{\recall}{Re}

\usepackage[capitalize,noabbrev]{cleveref}

\theoremstyle{plain}
\newtheorem{theorem}{Theorem}[section]
\newtheorem{proposition}[theorem]{Proposition}

\theoremstyle{definition}
\newtheorem{definition}[theorem]{Definition}

\theoremstyle{remark}

\definecolor{todocolor}{RGB}{210, 69, 69}

\begin{document}

\twocolumn[
\icmltitle{AttributionLab: Faithfulness of Feature Attribution Under Controllable Environments}



\icmlsetsymbol{equal}{*}

\begin{icmlauthorlist}
\icmlauthor{Yang Zhang}{equal,nus}
\icmlauthor{Yawei Li}{equal,lmu,mcml}
\icmlauthor{Hannah Brown}{nus}
\icmlauthor{Mina Rezaei}{lmu,mcml}
\icmlauthor{Bernd Bischl}{lmu,mcml}
\icmlauthor{Philip Torr}{ox}
\icmlauthor{Ashkan Khakzar}{ox}
\icmlauthor{Kenji Kawaguchi}{nus}
\end{icmlauthorlist}

\icmlaffiliation{nus}{National University of Singapore}
\icmlaffiliation{lmu}{LMU Munich}
\icmlaffiliation{mcml}{Munich Center for Machine Learning}
\icmlaffiliation{ox}{University of Oxford}

\icmlcorrespondingauthor{Yang Zhang}{yangzhang@u.nus.edu}
\icmlcorrespondingauthor{Yawei Li}{yawei.li@stat.uni-muenchen.de}

\icmlkeywords{Machine Learning, ICML}

\vskip 0.3in
]



\printAffiliationsAndNotice{\icmlEqualContribution} 

\begin{abstract}
Feature attribution explains neural network outputs by identifying relevant input features. The attribution has to be \emph{faithful}, meaning that the attributed features must mirror the input features that influence the output.
One recent trend to test faithfulness is to fit a model on designed data with known relevant features and then compare attributions with ground truth input features.
This idea assumes that the model learns to use \emph{all} and \emph{only} these designed features, for which there is no guarantee.
%
In this paper, we solve this issue by \emph{designing the network and manually setting its weights}, along with \emph{designing data}. 
%
The setup, AttributionLab, serves as a sanity check for faithfulness: If an attribution method is not faithful in a controlled environment, it can be unreliable in the wild. The environment is also a laboratory for controlled experiments by which we can analyze attribution methods and suggest improvements.

\end{abstract}

\section{Introduction}

\begin{figure*}[t]
    \centering
    \includegraphics[width=0.9\linewidth]{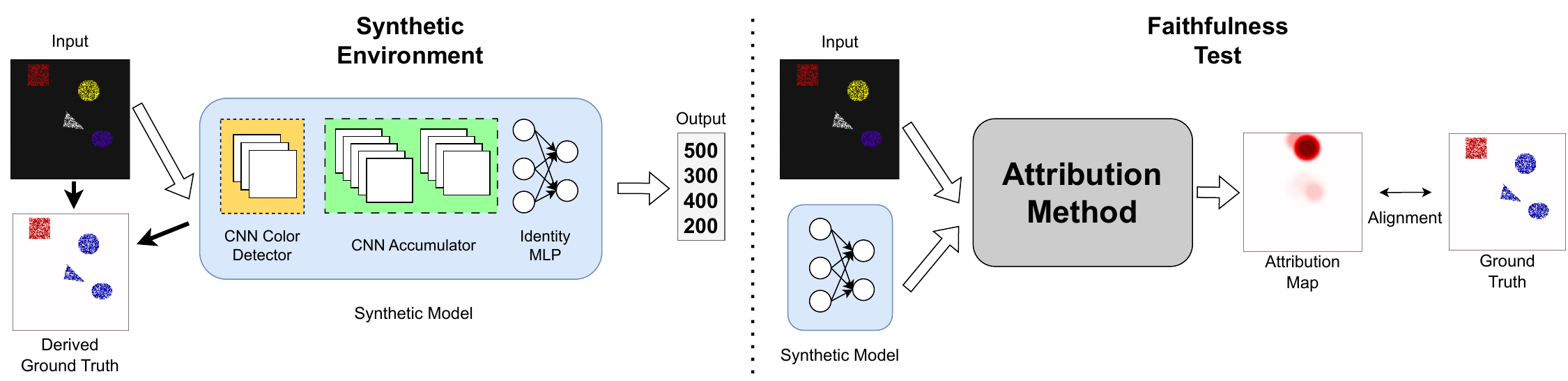}
    \caption{\textbf{Designing data and model} to set up a controllable environment for testing the faithfulness of attribution methods and analyzing their properties. To obtain the ground truth attribution, we explicitly design networks in tandem with inputs. The models follow conventional neural network designs and have sufficient complexity (shown in Table~\ref{table:model_summary}). More synthetic environments, including different modules (e.g., a modulo computer) and different tasks (e.g., regression) are in the appendix~\ref{sec:appendix:detail_model_design}. \textbf{The faithfulness test} performs a sanity check on attribution results in the synthetic setting with the ground truth attribution.
    }
    \label{fig:synthetic_environment}
\end{figure*}

\begin{figure*}[t]
    \centering
    \includegraphics[width=0.85\linewidth]{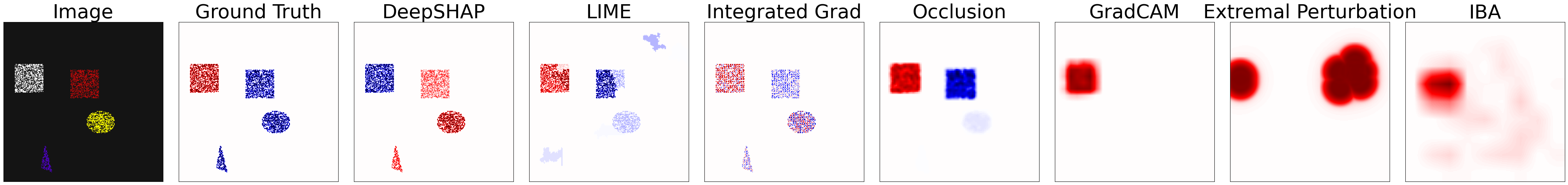}
    \caption{\textbf{Faithfulness test in AttributionLab}.
An attribution ($3^{rd}$ to the last image) is faithful if it is aligned with the Ground Truth attribution ($2^{nd}$ image) for a given input ($1^{st}$ image). When attributions are not faithful in this controlled environment, how can they be reliable in the wild? AttributionLab can serve as a sanity check for faithfulness and a tool to analyze current and future attribution methods.
In the following, we also analyze different factors that cause the misalignment. The exact setting for this figure is in Appendix \ref{sec:appendix:setting_for_visualization}.}
\label{fig:main:visualize_multi_color_sum}
\end{figure*}


Neural networks exhibit increasing capabilities as the scale of their design and their training data increases~\citep{caron2021emerging,wei2022emergent,oquab2023dinov2,openai2023gpt4,bubeck2023sparks}. These capabilities are achieved through the use of basic architectural blocks ~\citep{krizhevsky2012alexnet,vaswani2017attention}.
Though we know the architectural design of these networks and their computational graph explicitly, we do not have a human interpretable understanding of neural networks. 
One way to explain the neural network output is to \emph{identify important input features} for a single prediction, an explanation paradigm known as \emph{input feature attribution}.
There has been an ongoing quest for finding attribution methods to explain neural network functions~\citep{zeiler2014visualizing,selvaraju2017gradcam,shrikumar2017learning,sundararajan2017axiomatic,lundberg2017unified,schulz2020restricting,zhang2021fine}. However, one challenge remains:
How can we know whether the attributed features are aligned with input features influencing the output of the neural network? I.e., how do we know if an attribution is \emph{faithful}?
An attribution may seem reasonable to us, but the neural network may use other input features~\citep{ilyas2019adversarial}. Conversely, an attribution may seem unreasonable but be faithful and indeed reflect the features relevant to the neural networks.
In the presence of multiple differing attributions~\citep{krishna2022disagreement,khakzar2022explanations}, it is unclear which explanation to trust.

One recent trend to assess faithfulness is using synthetic data. Synthetic data are designed such that associations between features and labels are known to users~\citep{arras2022clevr,agarwal2023evaluating,zhou2022feature}. \emph{However, as we discuss in Section~\ref{sec:method:transparent_env}, there is no guarantee that the learning process will train the network to use the designed associations in the dataset}. Hence, the association learned by models can differ from the designed association in the synthetic data. Thus, evaluation based on the designed association is not guaranteed to reflect faithfulness. 

In scientific experiments with complex setups and multiple variables, it is typical to use a laboratory setting to conduct controlled tests.
Analogously, our work proposes a paradigm for providing a controlled laboratory environment. In this laboratory environment, \emph{both} the neural networks and the datasets are designed such that \emph{we know which features are relevant} to the network output. Thus, we obtain the \emph{ground truth attribution} in this synthetic environment. We leverage this information for the faithfulness test by measuring the alignment between the ground truth attribution and attribution maps (Section~\ref{sec:method:transparent_env} and~\ref{sec:method:design}).
If an attribution method is not faithful in the simulated setup, its performance in more complex scenarios is unreliable. A controlled environment can also be used to study the behavior of attribution methods under various circumstances by adjusting or ablating variables to simulate different scenarios (Section~\ref{sec:exp:eval_of_attr_methods}). 

%


\paragraph{Our contribution:}
(1) We argue with empirical evidence and theoretical supports that designing \emph{only} a synthetic dataset does not suffice to acquire ground truth attribution on a model, as a trained model can learn spurious features to achieve the task. (2) We propose a controllable environment consisting of a synthetic model and a synthetic dataset such that we obtain the ground truth attribution in this environment. A controllable environment can be used for our proposed faithfulness test as a form of sanity check for attribution methods and debugging attribution methods. We implement several controllable environments that cover various tasks. 
(3) We test mainstream attribution methods with our synthetic environment. Our experiment reveals several characteristics of these attribution methods. Based on the experiment results, we provide suggestions for future feature attribution research. (4) By simulating unseen data in AttributionLab we show that conventional perturbation analysis of faithfulness is unreliable.

\section{Faithfulness test in controllable settings}\label{sec:method:transparent_env}
We aim to establish an environment where we know exactly which input features are relevant to the output of a model. This laboratory setup allows for testing the faithfulness of feature attribution methods. Within this setup, we can identify which attributions are not aligned with the true attribution of the network and understand the sources of failures. Prior to detailing the setup, we underline the necessity of designing both the \emph{data} and the \emph{network} to obtain the ground truth attribution.

To realize the controlled setup for evaluating explanations, one might design a dataset where the associations between input features and output labels are known: e.g., a simple dataset of squares and circles. Indeed, prior approaches~\citep{arras2022clevr,zhou2022feature,schuessler2021two4two,agarwal2023evaluating} design synthetic datasets and train a model on them. The underlying assumption is that a model learns the intended association in the dataset once the model achieves peak performance on the synthetic dataset during the learning process. However, it is unclear whether the trained network uses the entire square, straight lines, or just corners to identify the square. How can we know if a network uses an entire square for decision-making in this setup? This phenomenon is formulated as the \emph{Rashomon effect}, which we rephrase in the following definition:

\paragraph{Rashomon effect~\citep{Breiman2001rashomon}}\label{def_rashomon_effect} \textit{(multiplicity of equally good models) There exist many models under the same hypothesis class that can achieve equally good accuracy but use different information for inference.}

\begin{figure}[t]
    \centering
    \includegraphics[width=\columnwidth]{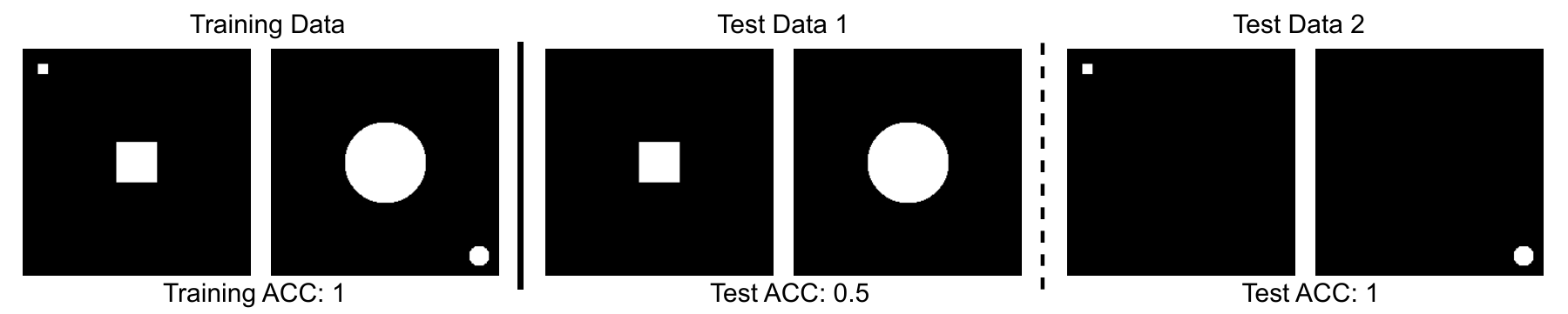}
    \caption{\textbf{Designing data is not enough.} Example on the neural networks not learning the designated ground truth features in the synthetic dataset. In this example, designed ground truth features are both objects in the center and on the edge. Even though the model can achieve $100\%$ accuracy, our test shows that the model only learns to use designed features at the corner and ignore the central ground truth features (more detail in Appendix~\ref{sec:appendix:trained_nn_failure}). 
    }
    \label{fig:trained_nn}
\end{figure}

The Rashomon effect states that it is generally invalid to assume that a model with $100\%$ accuracy on a dataset $\mathcal{D}$ is ensured to learn the true labeling process. In reality, depending on the data distribution, we can have trained models that learn to ignore some features relevant to the labels. For instance, the model can learn other associations (e.g. use corners to classify a square). 
Specifically, neural networks tend to be over-parametrized for their learning tasks, hence making it harder to learn the true labeling function \citep{ba2014deep,frankle2018the}. We provide empirical evidence for trained neural networks ignoring designed features (detailed setting in Appendix~\ref{sec:appendix:trained_nn_failure}). The result in Figure~\ref{fig:trained_nn} shows a neural network, which achieves $100\%$ training accuracy but learns to solely use partial ground truth features designed for the dataset (objects at the edge). 
In a nutshell, a model can learn to perform correctly on the dataset but is not guaranteed to learn the intended ground truth features in the data. Furthermore, studies~\citep{chen2019robust, singh2020attributional} demonstrate that attributions for a \emph{learned} model are very fragile to spurious correlations (e.g. objects at the edge in above example). \emph{Therefore, we manually design and set the weights of the neural network}.

\begin{figure*}[t]
    \centering
    \includegraphics[width=0.9\linewidth]{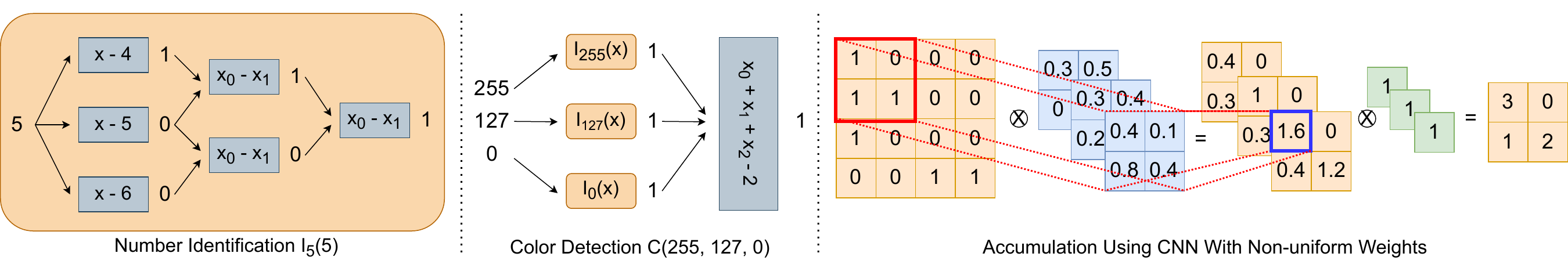}
    \caption{\textbf{Computational graph illustration} of our designed neural network modules. The left example shows a neural network of identifying number $5$, and the middle example shows a simple color detector for detecting RGB value $(255, 127, 0)$. In these two cases, blue boxes symbolize neurons, with their respective computations indicated within the box. ReLU activation is applied after each neuron, which is omitted in the figure. The right example demonstrates CNN operations to achieve accumulation using non-uniform kernel weights. More details can be found in Appendix~\ref{sec:appendix:detail_model_design}.}
    \label{fig:comp_graph}
\end{figure*}
By explicitly designing the data and the neural network according to specific properties, we know the ground truth features (features used by the network). 
%
Though designed neural networks can accurately provide ground truth attribution information, feature attribution evaluation only on designed neural networks might be biased. \emph{Hence, instead of using a synthetic environment for evaluation, we apply it for the faithfulness test as a form of sanity check for attribution methods.} 
Attribution methods are usually designed to be model agnostic, that is, they do not assume a target model and can work in general. We concretize this claim with the following definition. 
Given a model class $\mathcal{F} = \{f|f:\mathbb{R}^m \rightarrow \mathbb{R}^n\}$, a dataset $D=\{d: d\in\mathbb{R}^{m+n}\}$, an attribution method $\mathcal{A}: \mathcal{F}\times \{D\}\rightarrow \mathbb{R}^m$, and a faithfulness measurement $M: \mathbb{R}^m\times\mathcal{F}\times\{D\}\rightarrow\mathbb{R}$, we define $(\gamma, \mathcal{F})$-model agnostic faithfulness as follows:
\begin{definition}
    \label{def:faithfulness}
    \textbf{$(\gamma, \mathcal{F})$-model agnostic faithfulness: }An attribution method $\mathcal{A}$ is $(\gamma, \mathcal{F})$-model agnostic faithful, if 
    \begin{equation}
        \forall f \in \mathcal{F}, M(\mathcal{A}(f, D), f, D)\geq\gamma.
    \end{equation}
\end{definition}
An attribution method that has $(\gamma, \mathcal{F})$-model agnostic faithfulness should have faithfulness measurement results over threshold $\gamma$ for all models belonging to model class $\mathcal{F}$. This is a mild definition, as it is defined over a set of limited models, while attribution methods also usually claim to work on a specific set of models. In addition, this definition still allows the attribution method's faithfulness to vary in different models. However, $(\gamma, \mathcal{F})$-model agnostic faithfulness is hard to determine, as it would require validation through all possible models in the model class $\mathcal{F}$. Nevertheless, we can test if an attribution method is not $(\gamma, \mathcal{F})$-model agnostically faithful through proposed faithfulness test.

\paragraph{Faithfulness test: }An attribution method is not $(\gamma, \mathcal{F})$-model agnostic faithful, if 
\begin{equation}
    \exists f \in \mathcal{F}, M(\mathcal{A}(f, D), f, D)<\gamma.
\end{equation}

The faithfulness test only checks whether an attribution method is not model agnostic faithful, passing the test does not mean a method has model agnostic faithfulness. Nevertheless, this test provides an efficient and accurate means to identify underperforming attribution methods and ineffective settings for these methods. Previously, the difficulty lies in finding an accurate faithfulness measurement $M$. This is solved by our synthetic environment with designed $f$ and $D$. The main challenge remains in designing a model that can be considered to be in the model class $\mathcal{F}$. We discuss in the following section how we design synthetic models with conventional architecture and scale to satisfy this constraint.
\section{Design of data and neural network} \label{sec:method:design}
We propose a modular setup where each component performs specific tasks. For the designs to facilitate the faithfulness test on $(\epsilon, \mathcal{F})$-model agnostic faithfulness, we follow certain design principles. Firstly, the designs resemble real scenarios such as image classification using a convolutional neural network (Figure~\ref{fig:synthetic_environment}), to ensure that designed models belong to the desired model class. 
Furthermore, the design ensures that every ground-truth pixel is relevant and equally relevant. Specifically, with designed input data comprised of ground-truth (foreground) and baseline (background) pixels, designed neural networks have the following properties:
\begin{proposition}\label{def:sensitivity}
\textbf{(Sensitivity property)} 
The addition/removal of \textbf{any} ground-truth pixel to/from the background affects the output of the model.
\end{proposition}
\begin{proposition}\label{def:symmetry}
\textbf{(Symmetry property)} 
The addition/removal of any ground-truth pixel to/from the background \textbf{equally} affects the output of the model.
\end{proposition}
The Sensitivity property implies that in this designed model and dataset setup, \emph{every} ground truth pixel is relevant to the neural network output. The symmetry property implies that every ground truth pixel in the synthetic dataset should have \emph{equal} relevance to the output of the synthetic neural network.
In fact, these properties are aligned with the sensitivity and symmetry feature attribution axioms ~\citep{ManyShapleySundararajan2020,sundararajan2017axiomatic,lundberg2017unified}, which axiomatically define the relevance of a feature. 

As our main environment, we design a setup that resembles real image classification tasks. The task is to identify the dominant color (in terms of number of pixels) within an input image (see Figure ~\ref{fig:synthetic_environment}). The network first identifies predetermined colors and then counts the pixels for each. The designed dataset comprises color images, each containing $N_C$ patches of uniquely colored foreground objects and a distinct background color that is consistent across images, resulting in $N_C + 1$ colors in every image. The images are input to the model in RGB format. By treating each foreground color as a class, we formulate a multi-class task to predict the dominant color in an image. The designed classification model outputs softmax scores using logits for $N_C$ classes, where the logit of each class corresponds to the sum of pixels of the respective color. The softmax operation on the neural network output makes pixels of colors other than the target negatively contribute to the output of the target color (after softmax). Thus the ground truth will have both positively and negatively contributing features, enabling us to test whether an attribution method is able to discern between positive and negative contributions. The design of the model and the dataset also follow the properties in the following propositions (Proposition ~\ref{def:sensitivity} and ~\ref{def:symmetry}). \emph{Hence, we can infer that all pixels within the $N_C$ colored patches are relevant features}.

\paragraph{Simulating ``Unseen Data Effect"} It is common that trained neural networks have unintended outputs given inputs that are not from the training distribution. However, since the data and every module (and weights) are set up manually, we know the expected behavior of the network for any input. Specifically, the network exclusively counts the number of pixels of predetermined colors, and other input values (colors) do not affect the network output. Through a neural operation (explained below), we can upgrade the setup to simulate the behavior of trained neural networks given data not previously seen by the model during training. 

\subsection{Model design details}

The network is designed with an additional mode of behavior to imitate the behavior of trained neural networks. The default mode is that the model only responds to the predetermined colors and does not respond to other color values. The additional mode is that the model exhibits unpredictable behavior when given inputs that are not predetermined within the design (hence simulating the ``Unseen Data Effect"). To realize these effects, the model has the following components. The first component is a CNN color detector responsible for detecting target colors and simulating Unseen Data Effects. Its output has dimensions of $(N_C + N_R) \times H \times W$, where $N_C$ denotes the number of target classes and $N_R$ denotes the number of \emph{redundant} channels (the redundant channels are the neural implementation of ``Unseen Data Effect" simulation). For the first $N_C$ channels, the $i^{th}$ output map is the activation of the $i^{th}$ target color. We briefly explain how to implement color detection in neural networks. 
Firstly, we design a neural structure that can identify a specific integer number. For integer input, this structure can be defined as
\begin{equation}\label{eq:identify_number}
    \begin{aligned}
        I_N(x) &= \relu(I_{>N-1}(x) - I_{>N}(x)) \\
        \text{where } I_{>i}(x) &= \relu(\relu(x-i) - \relu(x-i-1)).
    \end{aligned}
\end{equation}
Given a color to be detected that is in RGB format $(R, G, B)$, where $R$, $G$, and $B$ are integers within $[0, 255]$. For a pixel with intensity $(r, g, b)$, the color detection mechanism shown in Figure~\ref{fig:comp_graph} is:
\begin{equation}
    C(r, g, b) = \relu(I_R(r) + I_G(g) + I_B(b) - 2).
\end{equation}
Here, the number identification functions $I_R$, $I_G$, and $I_B$ each detect a predefined component of an RGB value and are implemented as shown in Equation~\ref{eq:identify_number}. Hence, we have $C(r, g, b) = 1$ for $r=R, g=G,b=B$, and $C(r, g, b) = 0$ otherwise. 
The remaining $N_R$ redundant channels activate on any other colors not among the $N_C + 1$ colors. Specifically, if any pixel has a color that is neither a target color nor the background color, all $N_R$ redundant channels will activate at the position of this pixel. 
The activation mechanism of redundant channels is implemented as 
\begin{equation}
    R(r, g, b) = \relu( - \sum C_i(r, g, b) + 1). 
\end{equation}
Consequently, $R(r ,g ,b) = 1$ if all $C_i(r,g,b) = 0$, and $R(r ,g ,b) = 0$ if $C_i(r,g,b) = 1$ for any $i$.
Following the color detector, we have a CNN module that accumulates activation of the first $N_C$ channels respectively. We realize CNN accumulation in two settings, one with uniform weights in CNN, the other setting with non-uniform CNN weights. Figure~\ref{fig:comp_graph} illustrates the working principle of pixel accumulation using CNN with non-uniform weights (more details in Appendix~\ref{sec:appendix:accumulator}). The remaining $N_R$ redundant channels have random connections to the input of this CNN module. 
Therefore, the CNN accumulation module will have unexpected outputs if the input images contain any color that is not seen in the training dataset. Lastly, we have an MLP that performs identity mapping, as the output of the CNN module already provides the logits of each target color for normal inputs. 
The rationale behind adding an identity MLP module is to preserve the conventional model architecture in image classification, where a model is typically designed as a CNN followed by an MLP head.


\begin{table}[t]
    \centering
    \caption{
        Model summary in terms of the number of layers and parameters. Ours refers to \emph{one of} our designed models (detailed in Appendix~\ref{sec:app:model_summary_detail}). All designed models in AttributionLab have adequate parameters and layers to simulate conventional CNN architectures. }
    \begin{tabular}{c|c|c}
        \toprule
        Model & \# layers & \# parameters \\ 
        \midrule
        Ours & 17 & 675,033 \\ 
        \midrule
        ResNet-8~\citep{he2016deep} & 10 & 1,227,594 \\
        \bottomrule
    \end{tabular}
    \label{table:model_summary}
\end{table}

\paragraph{Other synthetic settings}
We provide \textbf{other synthetic settings} in Appendix~\ref{sec:appendix:single-color-modulo-setting}. The incentive for having additional synthetic settings is to cover more use cases, for instance, gray input images, regression models, and other model weight design schemes (e.g., modulo computation). 

\section{Faithfulness test of attribution methods}
\label{sec:exp:eval_of_attr_methods}

In this section, we deploy the designed environment to test the $(\gamma, \mathcal{F})$-model agnostic faithfulness of attribution methods and analyze their various aspects. 
We first visualize the attribution methods in AttributionLab in Figure~\ref{fig:main:visualize_multi_color_sum}. This figure reveals a lack of consensus among attribution methods, even in our synthetic environments. To quantitatively test the alignment between attribution maps and ground truth masks, we use precision, recall, and \Fone-score. However, since attribution maps usually have continuous pixel values, we adjust the metrics accordingly. Given a set of features $\mathcal{S}$ indexed by the index set $\mathcal{J} = \{j \in \mathbb{N}^{+} | 1 \leq j \leq |\mathcal{S}| \}$, let $a_j$ be the attribution value of the $j^{th}$ feature, and $g_j$ be the ground truth value of the $j^{th}$ feature, the (soft) precision is calculated as $\precision(\mathcal{S}) = \sum_{j \in \mathcal{J}}|a_j \cdot g_j|/\sum_{j \in \mathcal{J}}|a_j|$, while the recall is defined as $\recall(\mathcal{S}) = \sum_{j \in \mathcal{J}}|a_j \cdot g_j|/\sum_{j \in \mathcal{J}}|g_j|$. We normalize the attribution maps to $[-1, 1]$ to constrain the recall within the range of $[0, 1]$, as the ground truth masks are binary. Given the precision and recall, we can easily calculate the \Fone-score as well. In multi-class classification, signed ground truth information is available, outlining the positive and negative contributions of features to the target class. 
To test attribution methods with signed ground truth, we compute the precision, recall, and $F_1$ by comparing the positive attribution with positive ground truth features (GT), negative attribution with negative GT, respectively. Furthermore, we test the entire attribution map using an unsigned union of both positive and negative ground truth: the \emph{overall ground truth}. This test takes into consideration all features that contribute to decision-making, ignoring their direction of contribution. To measure $(\gamma, \mathcal{F})$-model agnostic faithfulness in \cref{def:faithfulness}, we employ the $F_1$-score as the faithfulness measurement. We use $\gamma = 0.5$ and consider $(0.5, \mathcal{F})$-model agnostic faithfulness as the criterion for our faithfulness test in the main text.

\begin{figure}[t]
    \centering
    \begin{subfigure}[t]{0.46\columnwidth}
         \includegraphics[width=\columnwidth]{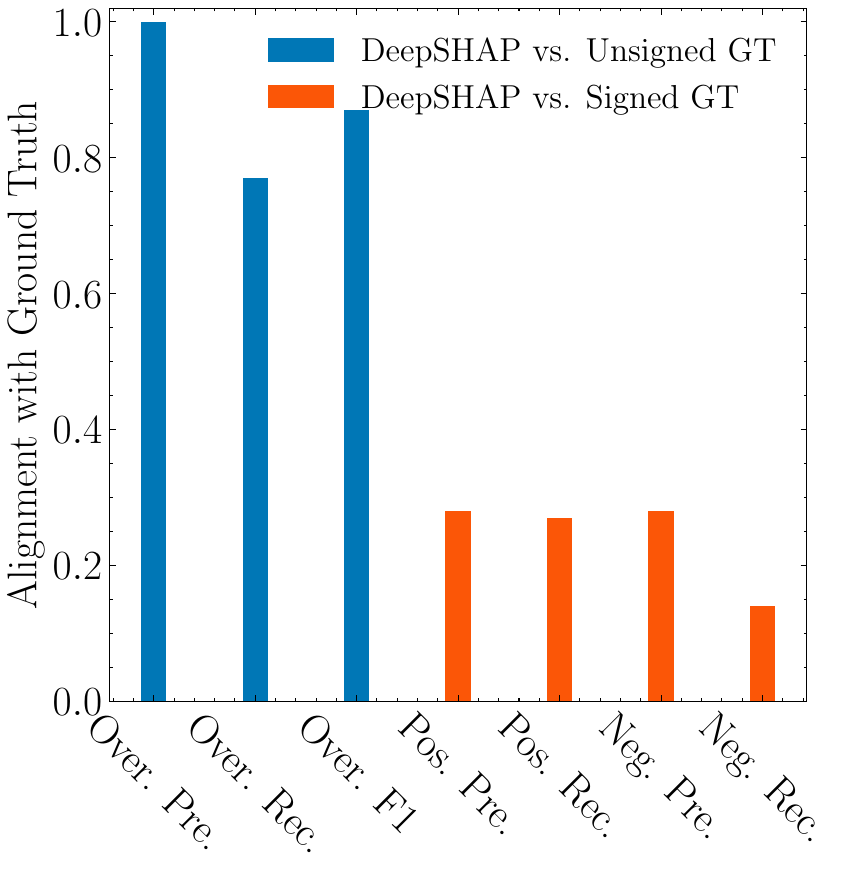}
        \caption{Faithfulness test} 
        \label{fig:main:deep_shap_barplot}
    \end{subfigure}
    \begin{subfigure}[t]{0.38\columnwidth}
        \centering
        \raisebox{2.0em}{\includegraphics[width=\columnwidth]{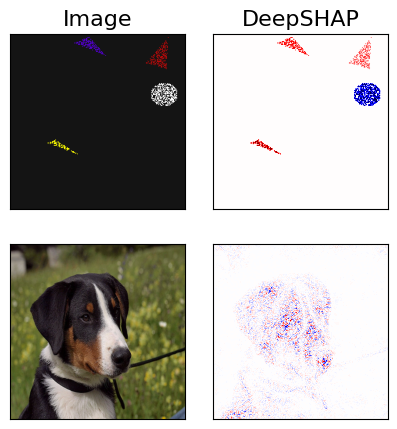}}
        \caption{Visual examples}
        \label{fig:main:deep_shap_visual}
    \end{subfigure}
    \caption{
        \textbf{Faithfulness test and visual examples of DeepSHAP.}
        (a) Faithfulness test of DeepSHAP. (b) Visual examples in synthetic and real-world environments. According to (a) and (b), DeepSHAP correctly highlights the foreground pixels. However, it assigns both positive and negative attribution to these pixels, even when they have similar colors and close spatial locations. 
    }
\end{figure}
\begin{figure}[t]
    \centering
    \begin{subfigure}[t]{0.4\columnwidth}
        \centering
        \includegraphics[width=\columnwidth]{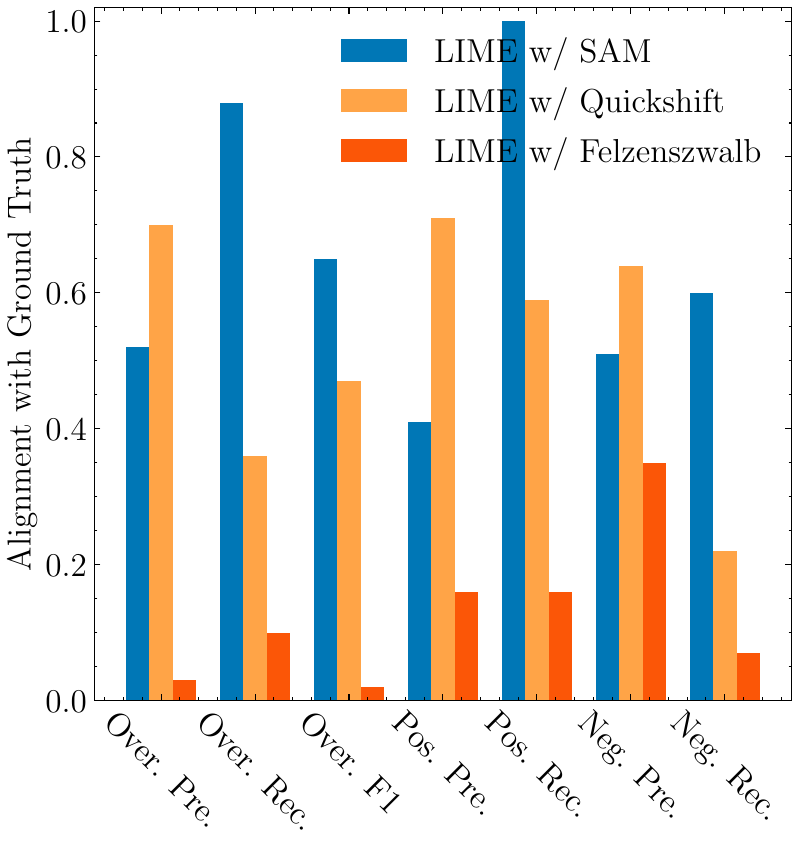}
        \caption{Faithfulness test}
        \label{fig:main:lime_qs_vs_fz_barplot}
    \end{subfigure}
    \begin{subfigure}[t]{0.59\columnwidth}
        \centering
        \raisebox{1.7em}{
        \includegraphics[width=\columnwidth]{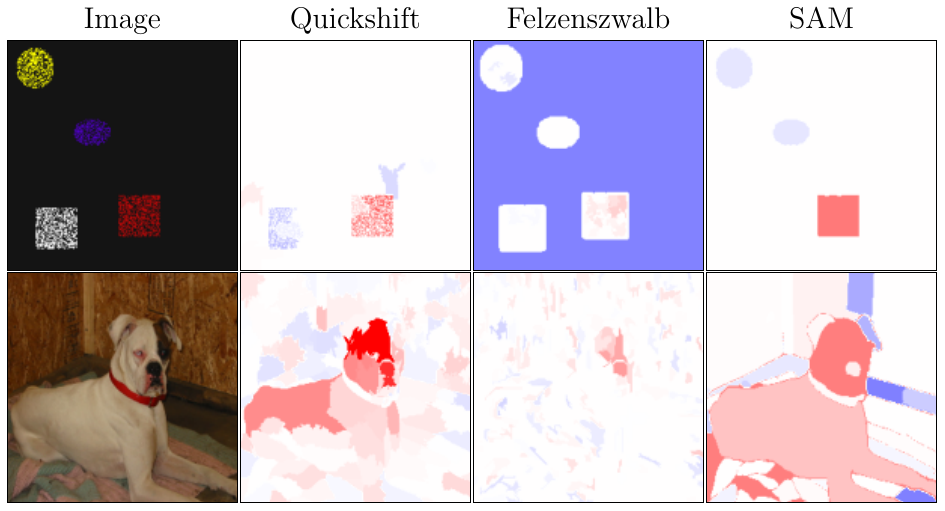}}
        \caption{Visual examples}
        \label{fig:main:lime_qs_vs_fz_visual}
    \end{subfigure}
    \caption{\textbf{LIME with diverse segmentation priors.} (a) Faithfulness test of LIME in AttributionLab. (b) Visual examples of LIME in synthetic and real-world environments.
    Results (a,b) show that the faithfulness of LIME depends on the segmentation prior. 
    }
\end{figure}
Additionally, we employ these attribution methods on an ImageNet~\citep{deng2009imagenet}-pretrained VGG16~\citep{simonyan2015a} to validate our findings from AttributionLab in real-world trained models. Specifically, we present visual samples from various attribution methods to affirm their consistent behaviors as observed in synthetic setups.

\subsection{DeepSHAP faithfulness analysis}


\begin{figure*}[t]
    \centering
    \begin{subfigure}[t]{0.28\textwidth}
        \centering
        \includegraphics[width=\columnwidth]{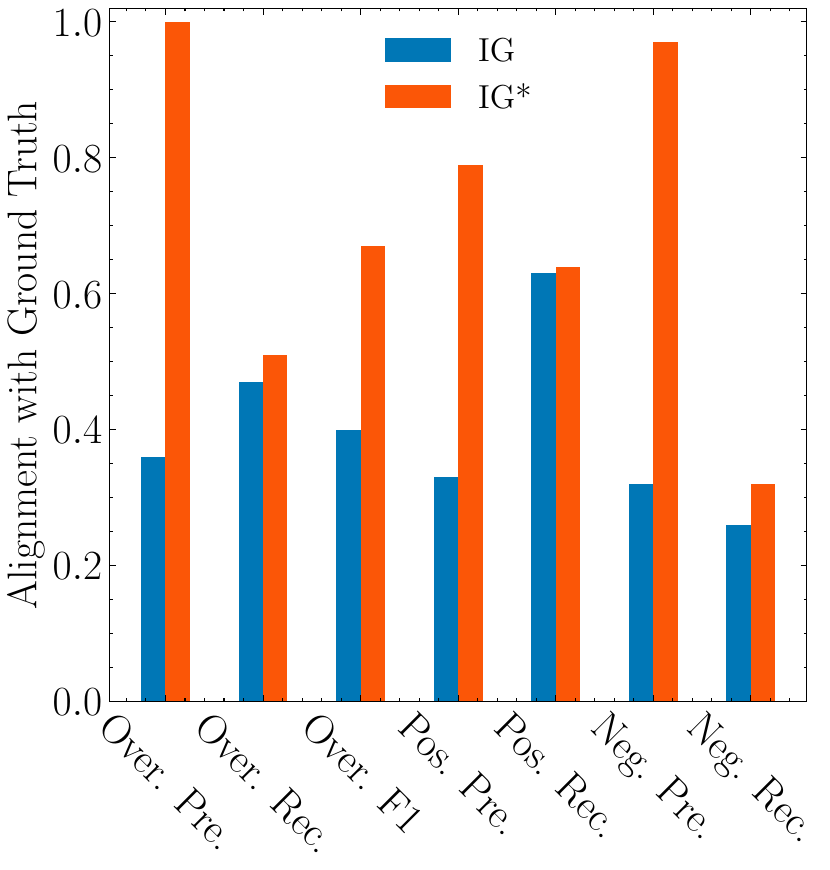}
        \caption{IG vs. $\text{IG}^*$}
        \label{fig:main:ig_baseline_ablation_barplot}
    \end{subfigure}
    \hfill
    \begin{subfigure}[t]{0.28\textwidth}
        \centering
        \includegraphics[width=\columnwidth]{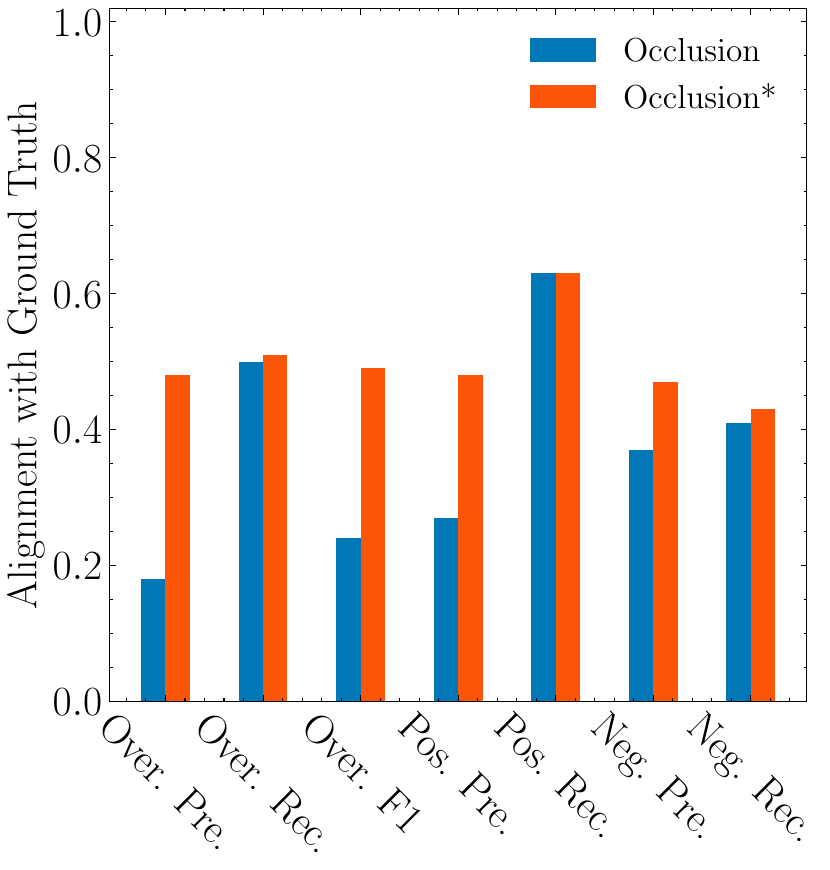}
        \caption{Occlusion vs. $\text{Occlusion}^*$}
        \label{fig:main:occ_baseline_ablation_barplot}
    \end{subfigure}
    \hfill
    \begin{subfigure}[t]{0.19\textwidth}
        \centering
        \raisebox{2.5em}{
            \includegraphics[width=\columnwidth]{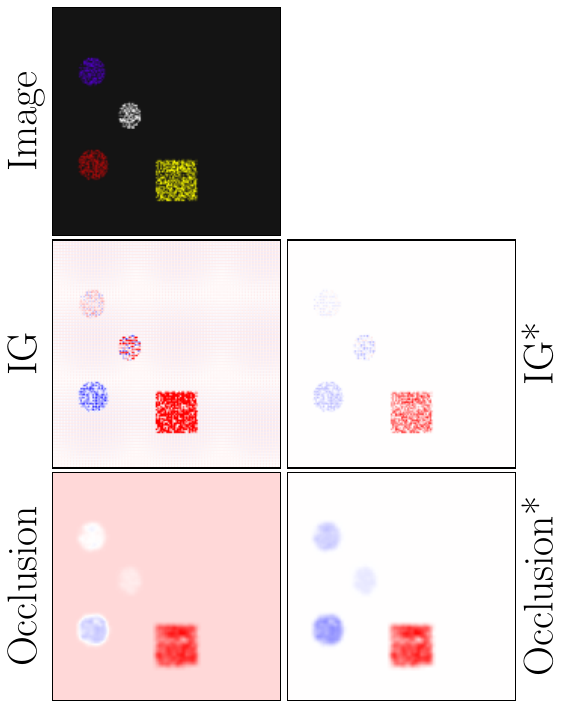}
        }
        \caption{Synthetic example}
        \label{fig:main:ig_occ_visual_multi_color_sum}
    \end{subfigure}
    \begin{subfigure}[t]{0.19\textwidth}
        \centering
        \raisebox{2.5em}{
            \includegraphics[width=\columnwidth]{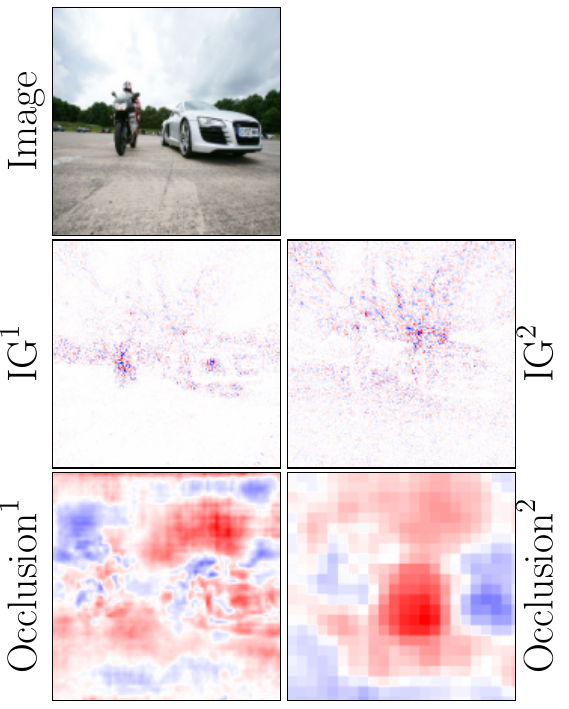}
        }
        \caption{Real example}
        \label{fig:main:ig_occ_visual_imagenet}
    \end{subfigure}
    \caption{\textbf{IG and Occlusion with diverse baselines}. (a-b) $\text{IG}^*$ and $\text{Occlusion}^*$ employing the ground truth baseline display substantial enhancement in faithfulness test. (c) Comparative visualization of attribution maps, created with and without the utilization of the ground truth baseline. (d) Attribution maps generated on ImageNet. The superscripts signify the use of distinct baselines. Notably, the attributions highlight different foreground objects when employing different baselines.}
    \label{fig:main:ig_occlusion_tests}
\end{figure*}

\textbf{— When considering all relevant features (discarding their sign), DeepSHAP passes the faithfulness test.}

DeepSHAP~\citep{lundberg2017unified} is designed to identify both positively and negatively contributing features. 
Figure~\ref{fig:main:deep_shap_barplot} shows the test result of DeepSHAP in the synthetic environment. 
The overall precision, recall, and \Fone-score reveal that DeepSHAP effectively locates the contributing features when we disregard the sign of attribution. Subsequently, DeepSHAP passes the faithfulness test with $\gamma=0.5$. However, the low precision and recall of positive and negative attribution suggest that DeepSHAP encounters difficulty in discerning whether a feature contributes positively or negatively to the target class. When we consider positive or negative attribution only, DeepSHAP fails the faithfulness test in both cases under the same $\gamma$. We further corroborated this issue in Figure~\ref{fig:main:deep_shap_visual}, which shows the phenomenon in both synthetic and real-world scenarios. In the ImageNet example shown in the image, we observe both positive and negative attribution on the \emph{foreground} object, especially on some pixels that are closely located and have similar colors. However, most of these highlighted pixels are likely to be relevant to the decision-making (though we cannot ascertain which pixels the model actually uses).
\emph{Our results suggest that DeepSHAP can be used to identify relevant features for the model, but is not effective in determining the direction of the contribution, although it is designed to do so. }

\subsection{LIME faithfulness analysis}\label{sec:exp:faithfulness_lime}

\textbf{— Using the right segmentation prior, LIME passes the faithfulness test.}

LIME~\citep{LIMEribeiro2016should} requires the input image to be segmented into superpixels, it then treats all pixels within a superpixel as a single feature. Consequently, the resulting attribution map can be influenced by the segmentation step. To investigate the impact of segmentation, we utilize Quickshift~\citep{vedaldi2008quick}, Felzenszwalb~\citep{felzenszwalb2004efficient}, and SAM~\citep{kirillov2023sam} segmentation algorithms and test the faithfulness of the resulting attributions. Figure~\ref{fig:main:lime_qs_vs_fz_barplot} reveals a noticeable difference between the outcomes derived from these segmentation techniques. Furthermore, LIME using Felzenszwalb fails to pass the faithfulness test with $\gamma=0.5$. Lime using Quickshift fails the faithfulness test with $\gamma=0.6$. However, LIME with SAM can pass the test for both values of $\gamma$. Figure~\ref{fig:main:lime_qs_vs_fz_visual} provides additional visual evidence of these differences in real-world scenarios. In LIME, the attribution of a superpixel conveys the aggregate predictive influence of that segment. This ``averaged'' attribution is then broadcast across all pixels within the superpixel. A more fine-grained attribution map requires a finer segmentation process. \emph{Therefore, a promising way to improve LIME is to find a better segmentation prior, that is, which pixels belong together and which ones are independent.}

\subsection{IG and Occlusion faithfulness analysis}
\textbf{— IG and Occlusion pass faithfulness tests \emph{only} with a proper baseline.}

Previous research~\citep{sturmfels2020visualizing} has revealed that IG~\citep{sundararajan2017axiomatic} is sensitive to the choice of baseline. While testing IG with a proper baseline remains a challenge in the real world, our controlled experimental setup provides the unique advantage of access to the true baseline. 
To study the effect of the baseline, we apply two settings. Here, $\text{IG}^{*}$ uses the predefined true baseline corresponding to background color $(20, 20, 20)$ in synthetic images, while IG employs a slightly incorrect baseline $(0, 0, 0)$ as default in many attribution applications. Figure~\ref{fig:main:ig_baseline_ablation_barplot} reveals a significant enhancement in the precision of $\text{IG}^{*}$ in our synthetic setting. Specifically, IG fails the faithfulness test with $\gamma=0.5$, while $\text{IG}^*$ passes. Analogous to $\text{IG}^*$, we introduce $\text{Occlusion}^*$ that employs the true baseline. Figure~\ref{fig:main:occ_baseline_ablation_barplot} also demonstrates a notably improved precision of $\text{Occlusion}^*$. Figure~\ref{fig:main:ig_occ_visual_multi_color_sum} and Figure~\ref{fig:main:ig_occ_visual_imagenet} further illustrate the sensitivity of these methods to the choice of baseline. 
Faithfulness test results show that IG and Occlusion are sensitive to the choice of baselines. In addition, they fail the faithfulness test given incorrect baselines. As methods like IG have adequate theoretical support, \emph{our findings underscore the importance of correct baselines for these methods}. Future research on these methods should focus on finding correct baselines to enhance their performance.

\begin{figure}[t]
    \centering
    \begin{subfigure}[t]{0.33\columnwidth}
         \includegraphics[width=\columnwidth]{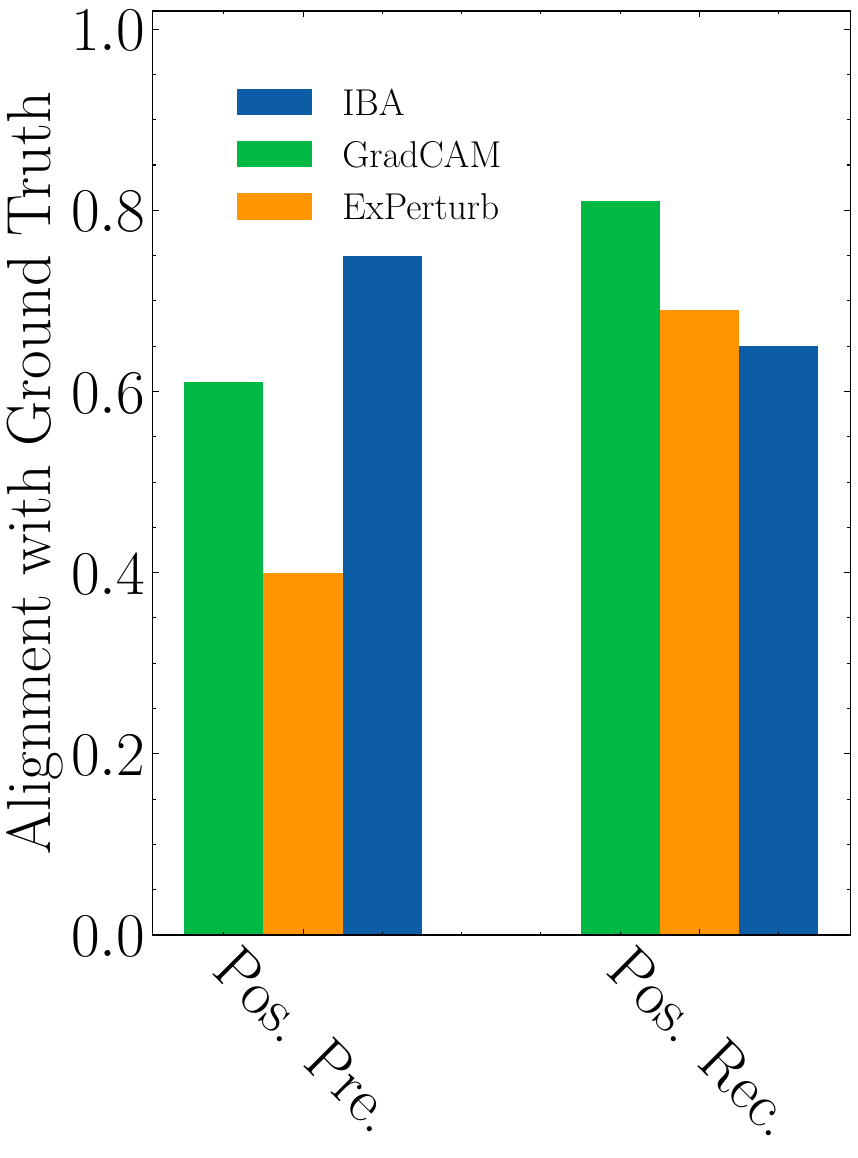}
        \caption{Faithfulness test} 
        \label{fig:main:four_methods_barplot}
    \end{subfigure}
    \begin{subfigure}[t]{0.66\columnwidth}
    \raisebox{1.9em}{
        \includegraphics[width=\columnwidth]{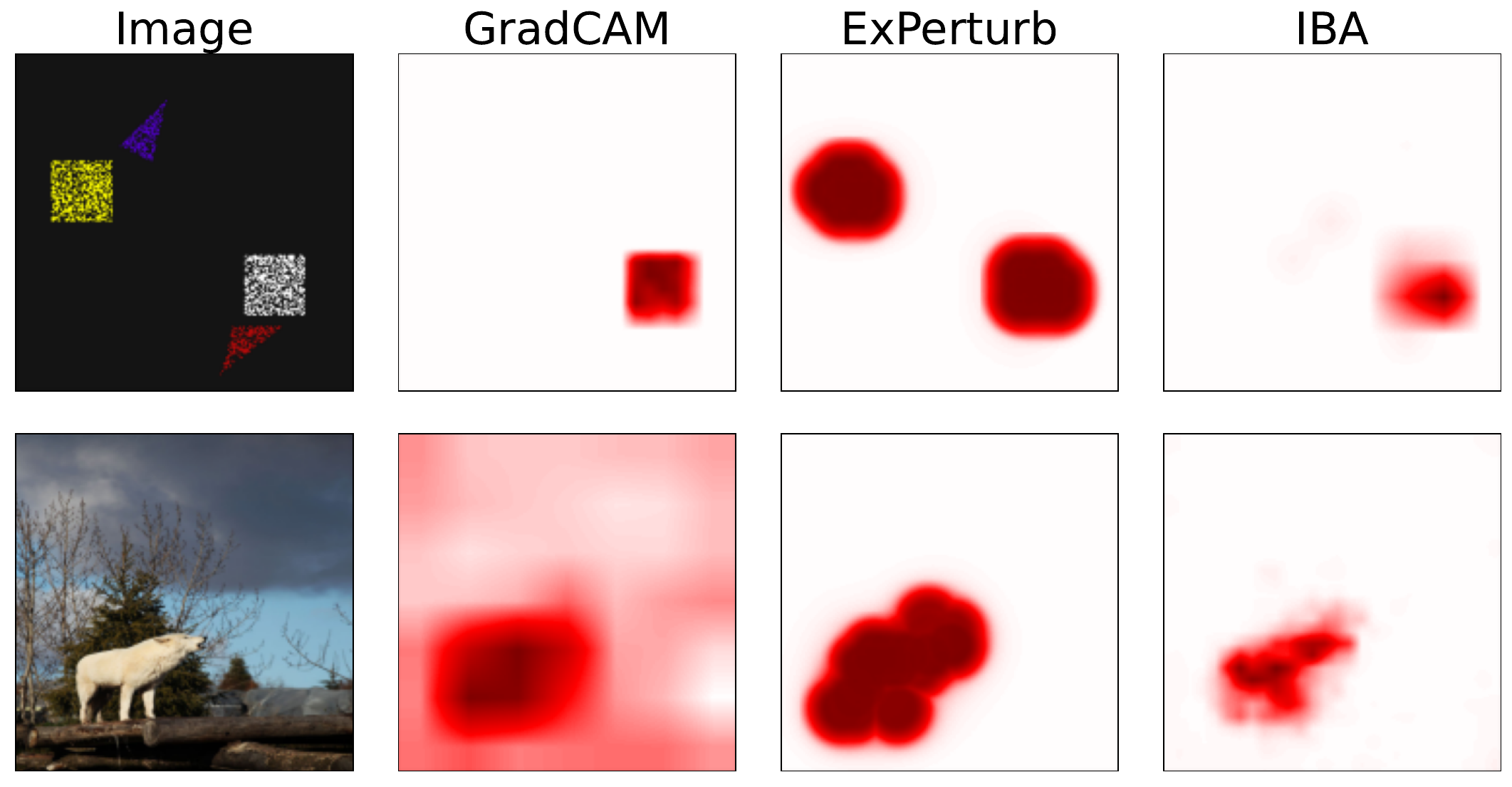}
        }
        \caption{AttributionLab and ImageNet samples}
        \label{fig:main:four_methods_visual}
    \end{subfigure}
    \caption{\textbf{Faithfulness test and visual examples for GradCAM, ExPerturb, and IBA}. All methods show promising recall in picking up positive features. ExPerturb has low precision (optimized mask can include other features). IBA is relatively the most faithful in identifying positive relevant features.}
    \label{fig:main:four_methods}
\end{figure}

\subsection{GradCAM, IBA and ExPerturb}
\textbf{— IBA and GradCAM pass the faithfulness test in identifying positive relevant features.}

These methods have different mechanisms: (GradCAM uses activations, ExPerturb uses mask optimization on input, and IBA masks the latent activations). However, they have two similar properties. First, they are not granular and are smooth (approximate) by design. Second, they are explicitly designed to provide positive contributions only. 
Thus, they are not capable of reflecting negatively relevant features. 
Therefore, we only analyze faithfulness to positively relevant features (Figure~\ref{fig:main:four_methods}), and to be fair to their granularity issue, we use smooth ground truth.
GradCAM~\citep{selvaraju2017gradcam} shows high recalls on identifying \emph{positively} contributing pixels. It resizes the attribution map to match the dimensions of the input. This resizing operation introduces blurriness (Figure~\ref{fig:main:four_methods_visual}), which consequently diminishes its precision. 
As for ExPerturb \cite{fong2019understanding}, the optimized mask also involves features that are not positively contributing to the output. Thus, the precision is low, though the recall for finding positive features is high. 
IBA \cite{schulz2020restricting} achieves high precision and recall similar to GradCAM. In conclusion, \emph{all three methods pass the faithfulness test for $(0.5, \mathcal{F})$-model agnostic faithfulness when we only consider positive contributions. }


\begin{figure}[t]
    \centering
    \begin{subfigure}[t]{0.45\columnwidth}
        \centering        \includegraphics[width=\columnwidth]{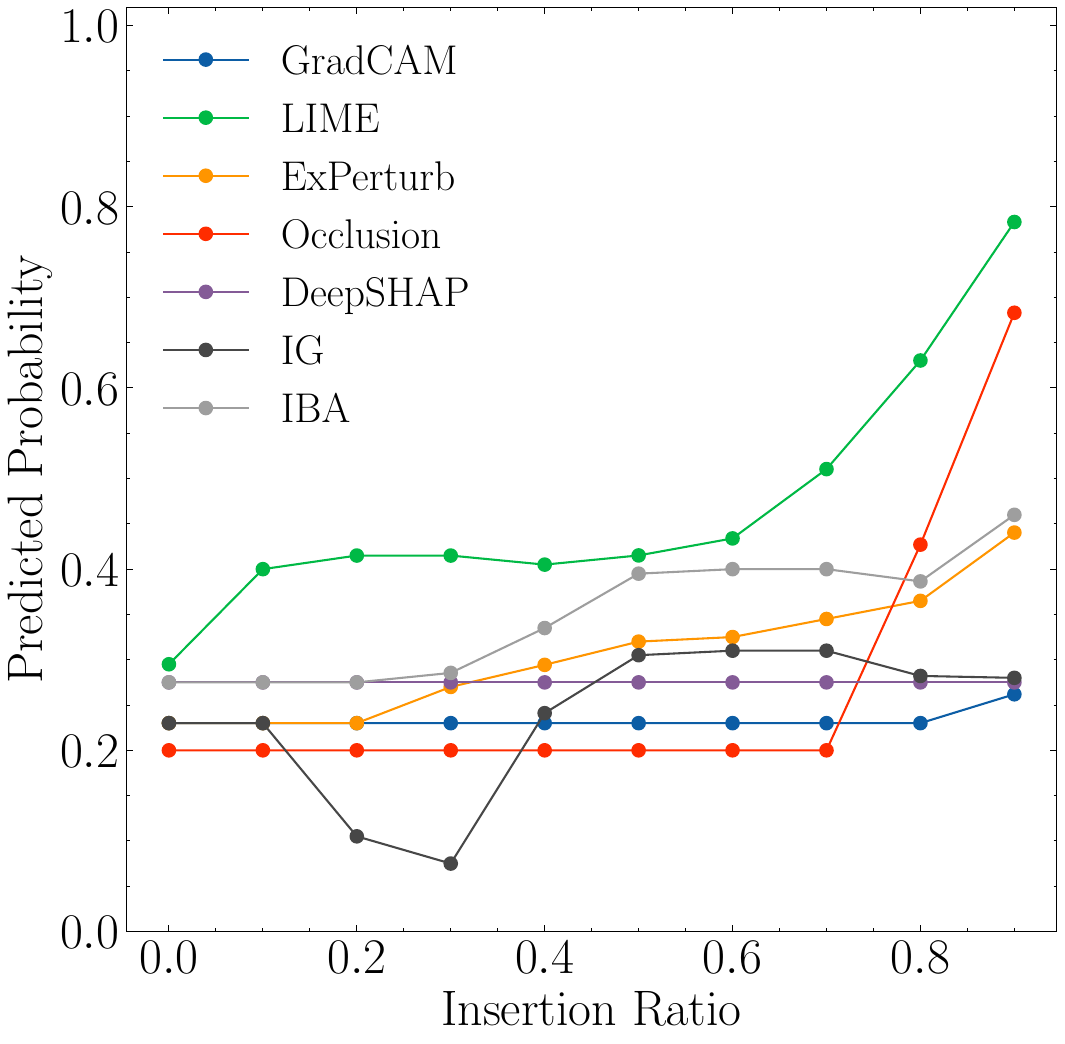}
        \caption{W/ Unseen Data Effect}
        \label{fig:main:ins_curves_multi_color_sum_non_uniform}
    \end{subfigure}
    \begin{subfigure}[t]{0.45\columnwidth}
        \centering
        \includegraphics[width=\columnwidth]{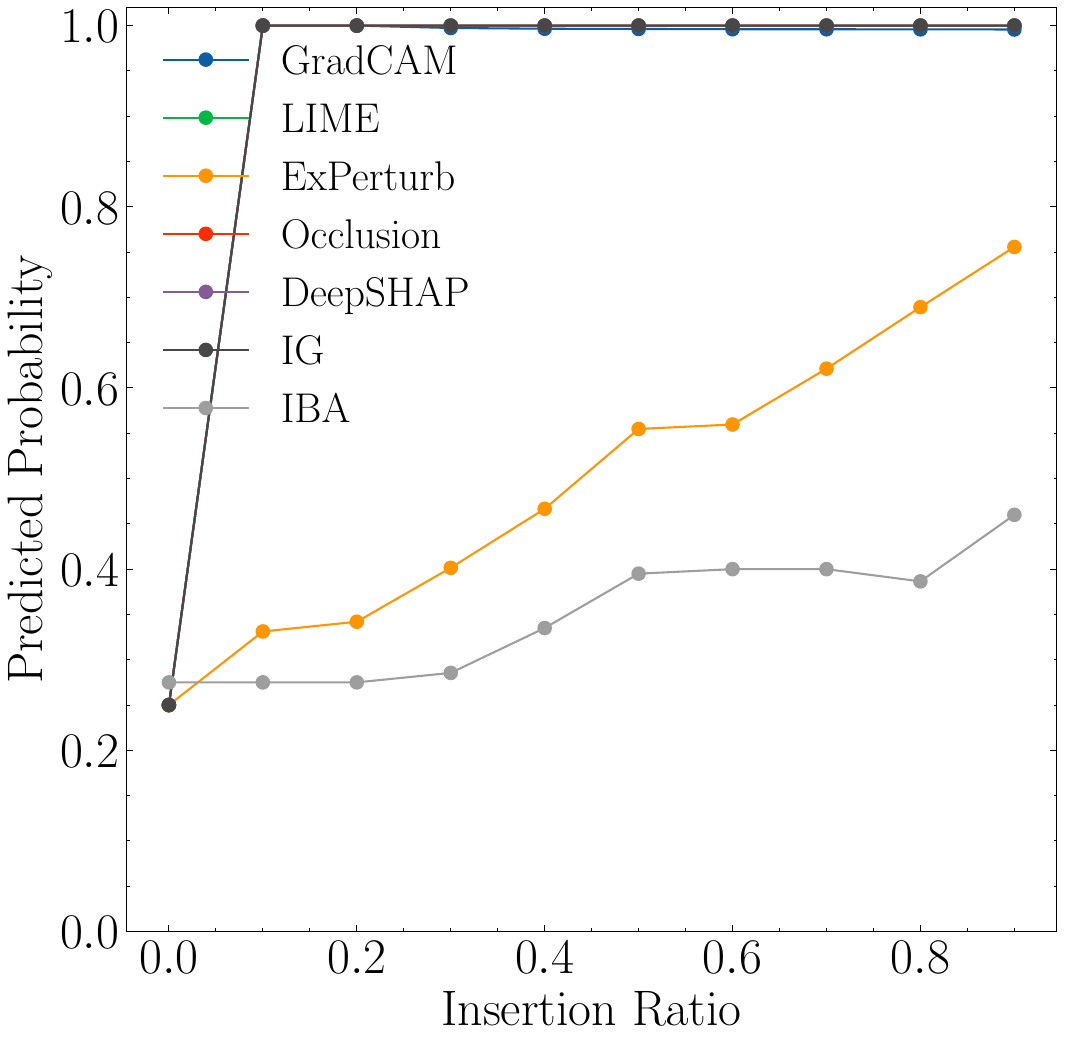}
        \caption{W/o Unseen Data Effect}
        \label{fig:main:ins_curves_multi_color_sum_wo_ood}
    \end{subfigure}
    \caption{\textbf{Sensitivity of perturbation-based evaluation metrics to Unseen Data Effect}. We test Insertion on the two AttributionLab setups. 
    In (b) several lines are overlapped. With the presence of the Unseen Data Effect, the evaluation result changes. Detailed results of Deletion and Sensitivity-N are in Appendix~\ref{sec:appendix:exp_multi_color_sum_non_uniform}. }
    \label{fig:main:ins_sens_n_curves}
\end{figure}

\section{Sanity check of faithfulness evaluations} \label{sec:exp:eval_of_eval_metrics}
\textbf{— We simulate the Unseen Data Effect in AttributionLab and observe that perturbation-based faithfulness evaluations are not reliable.}

There is another perspective through which faithfulness can be evaluated. One can perturb pixels in the image based on attribution values and observe how the output of the neural network changes. If the output change is aligned with the attribution values, then the attribution is considered faithful from this perspective. For instance, Insertion and Deletion~\citep{samek2016evaluating} progressively insert or remove pixels (Sensitivity-N~\citep{ancona2018towards} is in Appendix \ref{sec:appendix:related_work} and \ref{sec:appendix:sens_n_multi_color_sum_non_uniform}).

However, there is a fundamental issue lurking within this perspective on faithfulness. A significant output change can result from the network not having seen the new input resulting from perturbation, rather than the perturbed pixel being important. In the controlled environment of AttributionLab, we can observe how sensitive these evaluation metrics are to the Unseen Data Effect. That is when the pixels are perturbed during evaluation to another value for which the network is not trained, a scenario that happens easily in practice. 
We conduct experiments using these metrics on two AttributionLab setups to assess how this phenomenon (Unseen Data Effect) affects their performance. Figure~\ref{fig:main:ins_sens_n_curves} shows the order of feature attribution methods in terms of performance changes between the two scenarios for Insertion metrics. To confirm this observation, we compare the performance rankings (\Fone-score derived from \emph{positive} ground truth attribution) of attribution methods in these metrics with performance rankings established from the ground truth. Subsequently, we compute Spearman's rank correlation between the rankings, as demonstrated in Table~\ref{tab:main:spearmans_rank_correlation}. Additional experimental results are shown in Appendix~\ref{sec:appendix:exp_multi_color_sum_non_uniform}.
The first row of Table~\ref{tab:main:spearmans_rank_correlation} reveals that, in the presence of Unseen Data Effect, the metrics display substantially less consistency with the ground-truth-based evaluation. This inconsistency arises because Unseen Data Effect can result in unexpected predictions and inaccurate estimation of changes in model output. Hence, using model output scores as metrics may not accurately report the performance of attribution methods in the presence of Unseen Data Effect.

\begin{table}[t]
    \small
    \centering
    \caption{\textbf{Spearman's rank correlation}. The attribution method rankings are measured on perturbation-based metrics and the designed ground truth. In the presence of the Unseen Data Effect, these metrics show significant deviation from the ground truth.}
    \begin{tabular}{c c c c}
        \toprule
        Model has & \multirow{2}{*}{Insertion} & \multirow{2}{*}{Deletion} & \multirow{2}{*}{Sensitivity-N} \\
        Unseen Data Effect & & & \\
        \midrule
         Yes & 0.02 & 0.47 & 0.65 \\
        \midrule
         No & 0.42 & 0.61 & 0.81 \\
        \bottomrule
    \end{tabular}
    \label{tab:main:spearmans_rank_correlation}
\end{table}

\section{Related work}


Since the advent of deep neural networks~\citep{krizhevsky2012alexnet}, understanding how these complex models make predictions came under the spotlight \citep{simonyan2013deep}. One approach is \textbf{feature attribution}. Initial efforts to perform feature attribution focused on linear models~\citep{simonyan2013deep, bach2015pixel} and backpropagation~\citep{zeiler2014visualizing,springenberg2015striving}. Subsequently, more principled approaches emerged, such as backpropagating output differences with respect to a reference input~\citep{bach2015pixel, shrikumar2017learning} and axiomatic methods~\citep{sundararajan2017axiomatic, lundberg2017unified} inspired by the Shapley value~\citep{shapley1953value}. Meanwhile, intuitive approaches probing internal states~\citep{selvaraju2017gradcam, chattopadhay2018gradcampp, schulz2020restricting} and optimizing input masks were introduced~\citep{fong2019understanding,zhang2021fine}. With these advancements, \textbf{feature attribution evaluation} became a central question.
Literature has proposed \emph{sanity checks} testing whether attribution changes upon randomizing the network's weights~\citep{adebayo2018sanity} or if it is used on a different output class \citep{sixt2020explanations}. It also evaluates \emph{faithfulness} by analyzing networks's output when perturbing input features based on their relevance~\citep{samek2016evaluating, ancona2018towards, montavon2018methods, hooker2019benchmark, rong2022road}, or show theoretically if they are aligned with axioms~\citep{sundararajan2017axiomatic,lundberg2017unified}. Each reveals different aspects (Appendix~\ref{sec:appendix:related_work}). 
A recent trend in evaluating \emph{faithfulness} is designing datasets with known input-feature-output associations, enabling comparison between attribution and ground truth. \citet{arras2022clevr,zhou2022feature,khakzar2022explanations,agarwal2023evaluating}. However, these works do not guarantee that the network uses the intended ground truth associations. 


\section{Limitation}
We introduce an attribution faithfulness test by creating a synthetic environment, including a synthetic neural network and a synthetic dataset. 
Using our analysis in the controlled environment, we provide systematic suggestions for improving attribution methods.
While this approach offers an effective means for sanity check and troubleshooting of attribution methods, it has several limitations. 
Despite our efforts to design a neural network with conventional architectures and scales, our model does not fully resemble a trained neural network. 
Hence, designing other models that further imitate trained model behaviors can expand the test scenarios. 
The current study is limited to vision tasks, Convolutional Neural Networks (CNNs), and MLPs. Studies of transformers and other data modalities are not discussed in this work. With the addition of a self-attention module, transformers can be realized in follow-up research.

\section{Conclusion and future work}
This work proposes a novel, controlled laboratory setup for testing feature attribution explanations. 
The crux of our approach is the implementation of a paired design, i.e., manually programming the neural network and designing the dataset. We leverage our synthetic environment for a novel faithfulness test that efficiently and accurately checks whether an attribution method is \emph{not} model agnostically faithful. 
We test feature attribution explanations by simulating various conditions such as inappropriate baseline values for attribution methods and different segmentation masks as input of attribution methods, demonstrating their significant impact on attribution faithfulness. 
Our findings highlight the importance of correct settings of the existing attribution methods and encourage more exploration in finding adequate configurations than proposing new attribution methods.
We show that the Unseen Data Effect problem inherent in perturbation-based evaluations can negatively affect evaluation findings. 
Our proposed synthetic environment can empower future research by identifying not faithful methods and studying potential failure modes of attribution methods in a trustable, controlled environment. This could help researchers address issues before deploying methods in real-world settings, thus contributing to the development of more reliable and effective attribution methods.
\section*{Impact statements}
In this work, we introduce an evaluation framework designed specifically for feature attribution methods. 
There is a growing body of research focused on enhancing the explainability of large models in response to concerns surrounding their transparency. However, having wrong but positively affirming explanations, can have a large negative societal impact. Nonetheless, providing incorrect yet positively affirming explanations can potentially yield significant adverse societal consequences. A notable gap in the research community is the absence of reliable evaluation metrics, leading to various complications. For example, numerous feature attribution methods have been proposed that, in some cases, even yield inconsistent attribution results given identical inputs. Our approach guarantees its validity for evaluation purposes because of its fully synthetic nature. Our synthetic settings may appear primitive compared to real-world datasets and neural networks trained on such datasets. However, they provide a controlled laboratory environment, enabling a thorough examination of feature attribution methods prior to deployment. This facilitates a robust evaluation and refinement process for attribution methods. In this controlled environment we can identify the factors that render each attribution unfaithful, and provide a roadmap for large-scale analysis of these attributions.
\bibliography{icml24}

\begin{thebibliography}{53}
\providecommand{\natexlab}[1]{#1}
\providecommand{\url}[1]{\texttt{#1}}
\expandafter\ifx\csname urlstyle\endcsname\relax
  \providecommand{\doi}[1]{doi: #1}\else
  \providecommand{\doi}{doi: \begingroup \urlstyle{rm}\Url}\fi

\bibitem[Adebayo et~al.(2018)Adebayo, Gilmer, Muelly, Goodfellow, Hardt, and Kim]{adebayo2018sanity}
Adebayo, J., Gilmer, J., Muelly, M., Goodfellow, I., Hardt, M., and Kim, B.
\newblock Sanity checks for saliency maps.
\newblock \emph{Advances in neural information processing systems}, 31, 2018.

\bibitem[Agarwal et~al.(2023)Agarwal, Queen, Lakkaraju, and Zitnik]{agarwal2023evaluating}
Agarwal, C., Queen, O., Lakkaraju, H., and Zitnik, M.
\newblock Evaluating explainability for graph neural networks.
\newblock \emph{Scientific Data}, 10\penalty0 (1):\penalty0 144, 2023.

\bibitem[Ancona et~al.(2018)Ancona, Ceolini, Öztireli, and Gross]{ancona2018towards}
Ancona, M., Ceolini, E., Öztireli, C., and Gross, M.
\newblock Towards better understanding of gradient-based attribution methods for deep neural networks.
\newblock In \emph{International Conference on Learning Representations}, 2018.

\bibitem[Arras et~al.(2022)Arras, Osman, and Samek]{arras2022clevr}
Arras, L., Osman, A., and Samek, W.
\newblock Clevr-xai: A benchmark dataset for the ground truth evaluation of neural network explanations.
\newblock \emph{Information Fusion}, 81:\penalty0 14--40, 2022.
\newblock ISSN 1566-2535.
\newblock \doi{https://doi.org/10.1016/j.inffus.2021.11.008}.

\bibitem[Ba \& Caruana(2014)Ba and Caruana]{ba2014deep}
Ba, J. and Caruana, R.
\newblock Do deep nets really need to be deep?
\newblock \emph{Advances in neural information processing systems}, 27, 2014.

\bibitem[Bach et~al.(2015)Bach, Binder, Montavon, Klauschen, M{\"u}ller, and Samek]{bach2015pixel}
Bach, S., Binder, A., Montavon, G., Klauschen, F., M{\"u}ller, K.-R., and Samek, W.
\newblock On pixel-wise explanations for non-linear classifier decisions by layer-wise relevance propagation.
\newblock \emph{PloS one}, 10\penalty0 (7), 2015.

\bibitem[Breiman(2001)]{Breiman2001rashomon}
Breiman, L.
\newblock {Statistical Modeling: The Two Cultures (with comments and a rejoinder by the author)}.
\newblock \emph{Statistical Science}, 16\penalty0 (3):\penalty0 199 -- 231, 2001.
\newblock \doi{10.1214/ss/1009213726}.
\newblock URL \url{https://doi.org/10.1214/ss/1009213726}.

\bibitem[Bubeck et~al.(2023)Bubeck, Chandrasekaran, Eldan, Gehrke, Horvitz, Kamar, Lee, Lee, Li, Lundberg, et~al.]{bubeck2023sparks}
Bubeck, S., Chandrasekaran, V., Eldan, R., Gehrke, J., Horvitz, E., Kamar, E., Lee, P., Lee, Y.~T., Li, Y., Lundberg, S., et~al.
\newblock Sparks of artificial general intelligence: Early experiments with gpt-4.
\newblock \emph{arXiv preprint arXiv:2303.12712}, 2023.

\bibitem[Caron et~al.(2021)Caron, Touvron, Misra, J{\'e}gou, Mairal, Bojanowski, and Joulin]{caron2021emerging}
Caron, M., Touvron, H., Misra, I., J{\'e}gou, H., Mairal, J., Bojanowski, P., and Joulin, A.
\newblock Emerging properties in self-supervised vision transformers.
\newblock In \emph{Proceedings of the IEEE/CVF international conference on computer vision}, pp.\  9650--9660, 2021.

\bibitem[Chattopadhay et~al.(2018)Chattopadhay, Sarkar, Howlader, and Balasubramanian]{chattopadhay2018gradcampp}
Chattopadhay, A., Sarkar, A., Howlader, P., and Balasubramanian, V.~N.
\newblock Grad-cam++: Generalized gradient-based visual explanations for deep convolutional networks.
\newblock In \emph{2018 IEEE Winter Conference on Applications of Computer Vision (WACV)}, pp.\  839--847, 2018.
\newblock \doi{10.1109/WACV.2018.00097}.

\bibitem[Chen et~al.(2019)Chen, Wu, Rastogi, Liang, and Jha]{chen2019robust}
Chen, J., Wu, X., Rastogi, V., Liang, Y., and Jha, S.
\newblock Robust attribution regularization.
\newblock \emph{Advances in Neural Information Processing Systems}, 32, 2019.

\bibitem[Deng et~al.(2009)Deng, Dong, Socher, Li, Li, and Fei-Fei]{deng2009imagenet}
Deng, J., Dong, W., Socher, R., Li, L.-J., Li, K., and Fei-Fei, L.
\newblock Imagenet: A large-scale hierarchical image database.
\newblock In \emph{2009 IEEE Conference on Computer Vision and Pattern Recognition}, pp.\  248--255, 2009.
\newblock \doi{10.1109/CVPR.2009.5206848}.

\bibitem[Dosovitskiy et~al.(2020)Dosovitskiy, Beyer, Kolesnikov, Weissenborn, Zhai, Unterthiner, Dehghani, Minderer, Heigold, Gelly, et~al.]{dosovitskiy2020image}
Dosovitskiy, A., Beyer, L., Kolesnikov, A., Weissenborn, D., Zhai, X., Unterthiner, T., Dehghani, M., Minderer, M., Heigold, G., Gelly, S., et~al.
\newblock An image is worth 16x16 words: Transformers for image recognition at scale.
\newblock In \emph{International Conference on Learning Representations}, 2020.

\bibitem[Felzenszwalb \& Huttenlocher(2004)Felzenszwalb and Huttenlocher]{felzenszwalb2004efficient}
Felzenszwalb, P.~F. and Huttenlocher, D.~P.
\newblock Efficient graph-based image segmentation.
\newblock \emph{International journal of computer vision}, 59:\penalty0 167--181, 2004.

\bibitem[Fong et~al.(2019)Fong, Patrick, and Vedaldi]{fong2019understanding}
Fong, R., Patrick, M., and Vedaldi, A.
\newblock Understanding deep networks via extremal perturbations and smooth masks.
\newblock In \emph{Proceedings of the IEEE/CVF international conference on computer vision}, pp.\  2950--2958, 2019.

\bibitem[Frankle \& Carbin(2019)Frankle and Carbin]{frankle2018the}
Frankle, J. and Carbin, M.
\newblock The lottery ticket hypothesis: Finding sparse, trainable neural networks.
\newblock In \emph{International Conference on Learning Representations}, 2019.
\newblock URL \url{https://openreview.net/forum?id=rJl-b3RcF7}.

\bibitem[He et~al.(2016)He, Zhang, Ren, and Sun]{he2016deep}
He, K., Zhang, X., Ren, S., and Sun, J.
\newblock Deep residual learning for image recognition.
\newblock In \emph{Proceedings of the IEEE conference on computer vision and pattern recognition}, pp.\  770--778, 2016.

\bibitem[Hooker et~al.(2019)Hooker, Erhan, Kindermans, and Kim]{hooker2019benchmark}
Hooker, S., Erhan, D., Kindermans, P.-J., and Kim, B.
\newblock A benchmark for interpretability methods in deep neural networks.
\newblock \emph{Advances in neural information processing systems}, 32, 2019.

\bibitem[Ilyas et~al.(2019)Ilyas, Santurkar, Tsipras, Engstrom, Tran, and Madry]{ilyas2019adversarial}
Ilyas, A., Santurkar, S., Tsipras, D., Engstrom, L., Tran, B., and Madry, A.
\newblock Adversarial examples are not bugs, they are features.
\newblock \emph{Advances in neural information processing systems}, 32, 2019.

\bibitem[Khakzar et~al.(2021)Khakzar, Baselizadeh, Khanduja, Rupprecht, Kim, and Navab]{khakzar2021neural}
Khakzar, A., Baselizadeh, S., Khanduja, S., Rupprecht, C., Kim, S.~T., and Navab, N.
\newblock Neural response interpretation through the lens of critical pathways.
\newblock In \emph{Proceedings of the IEEE/CVF Conference on Computer Vision and Pattern Recognition}, pp.\  13528--13538, 2021.

\bibitem[Khakzar et~al.(2022)Khakzar, Khorsandi, Nobahari, and Navab]{khakzar2022explanations}
Khakzar, A., Khorsandi, P., Nobahari, R., and Navab, N.
\newblock Do explanations explain? model knows best.
\newblock In \emph{Proceedings of the IEEE/CVF Conference on Computer Vision and Pattern Recognition}, pp.\  10244--10253, 2022.

\bibitem[Kirillov et~al.(2023)Kirillov, Mintun, Ravi, Mao, Rolland, Gustafson, Xiao, Whitehead, Berg, Lo, Doll{\'a}r, and Girshick]{kirillov2023sam}
Kirillov, A., Mintun, E., Ravi, N., Mao, H., Rolland, C., Gustafson, L., Xiao, T., Whitehead, S., Berg, A.~C., Lo, W.-Y., Doll{\'a}r, P., and Girshick, R.
\newblock Segment anything.
\newblock \emph{arXiv:2304.02643}, 2023.

\bibitem[Krishna et~al.(2022)Krishna, Han, Gu, Pombra, Jabbari, Wu, and Lakkaraju]{krishna2022disagreement}
Krishna, S., Han, T., Gu, A., Pombra, J., Jabbari, S., Wu, S., and Lakkaraju, H.
\newblock The disagreement problem in explainable machine learning: A practitioner's perspective, 2022.

\bibitem[Krizhevsky et~al.(2012)Krizhevsky, Sutskever, and Hinton]{krizhevsky2012alexnet}
Krizhevsky, A., Sutskever, I., and Hinton, G.~E.
\newblock Imagenet classification with deep convolutional neural networks.
\newblock In \emph{Advances in Neural Information Processing Systems}, volume~25, 2012.

\bibitem[Lundberg \& Lee(2017)Lundberg and Lee]{lundberg2017unified}
Lundberg, S.~M. and Lee, S.-I.
\newblock A unified approach to interpreting model predictions.
\newblock \emph{Advances in neural information processing systems}, 30, 2017.

\bibitem[Montavon et~al.(2018)Montavon, Samek, and M{\"u}ller]{montavon2018methods}
Montavon, G., Samek, W., and M{\"u}ller, K.-R.
\newblock Methods for interpreting and understanding deep neural networks.
\newblock \emph{Digital signal processing}, 73:\penalty0 1--15, 2018.

\bibitem[OpenAI(2023)]{openai2023gpt4}
OpenAI.
\newblock Gpt-4 technical report, 2023.

\bibitem[Oquab et~al.(2023)Oquab, Darcet, Moutakanni, Vo, Szafraniec, Khalidov, Fernandez, Haziza, Massa, El-Nouby, et~al.]{oquab2023dinov2}
Oquab, M., Darcet, T., Moutakanni, T., Vo, H., Szafraniec, M., Khalidov, V., Fernandez, P., Haziza, D., Massa, F., El-Nouby, A., et~al.
\newblock Dinov2: Learning robust visual features without supervision.
\newblock \emph{arXiv preprint arXiv:2304.07193}, 2023.

\bibitem[Ribeiro et~al.(2016)Ribeiro, Singh, and Guestrin]{LIMEribeiro2016should}
Ribeiro, M.~T., Singh, S., and Guestrin, C.
\newblock " why should i trust you?" explaining the predictions of any classifier.
\newblock In \emph{Proceedings of the 22nd ACM SIGKDD international conference on knowledge discovery and data mining}, pp.\  1135--1144, 2016.

\bibitem[Rong et~al.(2022)Rong, Leemann, Borisov, Kasneci, and Kasneci]{rong2022road}
Rong, Y., Leemann, T., Borisov, V., Kasneci, G., and Kasneci, E.
\newblock A consistent and efficient evaluation strategy for attribution methods.
\newblock In \emph{International Conference on Machine Learning}, pp.\  18770--18795. PMLR, 2022.

\bibitem[Samek et~al.(2016)Samek, Binder, Montavon, Lapuschkin, and M{\"u}ller]{samek2016evaluating}
Samek, W., Binder, A., Montavon, G., Lapuschkin, S., and M{\"u}ller, K.-R.
\newblock Evaluating the visualization of what a deep neural network has learned.
\newblock \emph{IEEE transactions on neural networks and learning systems}, 28\penalty0 (11):\penalty0 2660--2673, 2016.

\bibitem[Schuessler et~al.(2021)Schuessler, Weiß, and Sixt]{schuessler2021two4two}
Schuessler, M., Weiß, P., and Sixt, L.
\newblock Two4two: Evaluating interpretable machine learning - a synthetic dataset for controlled experiments, 2021.

\bibitem[Schulz et~al.(2020)Schulz, Sixt, Tombari, and Landgraf]{schulz2020restricting}
Schulz, K., Sixt, L., Tombari, F., and Landgraf, T.
\newblock Restricting the flow: Information bottlenecks for attribution.
\newblock In \emph{International Conference on Learning Representations}, 2020.

\bibitem[Selvaraju et~al.(2017)Selvaraju, Cogswell, Das, Vedantam, Parikh, and Batra]{selvaraju2017gradcam}
Selvaraju, R.~R., Cogswell, M., Das, A., Vedantam, R., Parikh, D., and Batra, D.
\newblock Grad-cam: Visual explanations from deep networks via gradient-based localization.
\newblock In \emph{2017 IEEE International Conference on Computer Vision (ICCV)}, pp.\  618--626, 2017.
\newblock \doi{10.1109/ICCV.2017.74}.

\bibitem[Shapley(1953)]{shapley1953value}
Shapley, L.~S.
\newblock A value for n-person games.
\newblock \emph{Contributions to the Theory of Games}, 2\penalty0 (28):\penalty0 307--317, 1953.

\bibitem[Shrikumar et~al.(2017)Shrikumar, Greenside, and Kundaje]{shrikumar2017learning}
Shrikumar, A., Greenside, P., and Kundaje, A.
\newblock Learning important features through propagating activation differences.
\newblock In \emph{International conference on machine learning}, pp.\  3145--3153. PMLR, 2017.

\bibitem[Simonyan \& Zisserman(2015)Simonyan and Zisserman]{simonyan2015a}
Simonyan, K. and Zisserman, A.
\newblock Very deep convolutional networks for large-scale image recognition.
\newblock In \emph{3rd International Conference on Learning Representations (ICLR 2015)}, pp.\  1--14, 2015.

\bibitem[Simonyan et~al.(2013)Simonyan, Vedaldi, and Zisserman]{simonyan2013deep}
Simonyan, K., Vedaldi, A., and Zisserman, A.
\newblock Deep inside convolutional networks: Visualising image classification models and saliency maps.
\newblock \emph{arXiv preprint arXiv:1312.6034}, 2013.

\bibitem[Singh et~al.(2020)Singh, Kumari, Mangla, Sinha, Balasubramanian, and Krishnamurthy]{singh2020attributional}
Singh, M., Kumari, N., Mangla, P., Sinha, A., Balasubramanian, V.~N., and Krishnamurthy, B.
\newblock Attributional robustness training using input-gradient spatial alignment.
\newblock In \emph{European Conference on Computer Vision}, pp.\  515--533. Springer, 2020.

\bibitem[Sixt et~al.(2020)Sixt, Granz, and Landgraf]{sixt2020explanations}
Sixt, L., Granz, M., and Landgraf, T.
\newblock When explanations lie: Why many modified bp attributions fail.
\newblock In \emph{International Conference on Machine Learning}, pp.\  9046--9057. PMLR, 2020.

\bibitem[Smilkov et~al.(2017)Smilkov, Thorat, Kim, Viégas, and Wattenberg]{smilkov2017smoothgrad}
Smilkov, D., Thorat, N., Kim, B., Viégas, F., and Wattenberg, M.
\newblock Smoothgrad: removing noise by adding noise, 2017.

\bibitem[Springenberg et~al.(2015)Springenberg, Dosovitskiy, Brox, and Riedmiller]{springenberg2015striving}
Springenberg, J., Dosovitskiy, A., Brox, T., and Riedmiller, M.
\newblock Striving for simplicity: The all convolutional net.
\newblock In \emph{ICLR (workshop track)}, 2015.

\bibitem[Sturmfels et~al.(2020)Sturmfels, Lundberg, and Lee]{sturmfels2020visualizing}
Sturmfels, P., Lundberg, S., and Lee, S.-I.
\newblock Visualizing the impact of feature attribution baselines.
\newblock \emph{Distill}, 2020.
\newblock \doi{10.23915/distill.00022}.
\newblock https://distill.pub/2020/attribution-baselines.

\bibitem[Sundararajan \& Najmi(2020)Sundararajan and Najmi]{ManyShapleySundararajan2020}
Sundararajan, M. and Najmi, A.
\newblock The many shapley values for model explanation.
\newblock \emph{37th International Conference on Machine Learning, ICML 2020}, 2020.

\bibitem[Sundararajan et~al.(2017)Sundararajan, Taly, and Yan]{sundararajan2017axiomatic}
Sundararajan, M., Taly, A., and Yan, Q.
\newblock Axiomatic attribution for deep networks.
\newblock In \emph{International conference on machine learning}, pp.\  3319--3328. PMLR, 2017.

\bibitem[Vaswani et~al.(2017)Vaswani, Shazeer, Parmar, Uszkoreit, Jones, Gomez, Kaiser, and Polosukhin]{vaswani2017attention}
Vaswani, A., Shazeer, N., Parmar, N., Uszkoreit, J., Jones, L., Gomez, A.~N., Kaiser, {\L}., and Polosukhin, I.
\newblock Attention is all you need.
\newblock \emph{Advances in neural information processing systems}, 30, 2017.

\bibitem[Vedaldi \& Soatto(2008)Vedaldi and Soatto]{vedaldi2008quick}
Vedaldi, A. and Soatto, S.
\newblock Quick shift and kernel methods for mode seeking.
\newblock In \emph{Computer Vision--ECCV 2008: 10th European Conference on Computer Vision, Marseille, France, October 12-18, 2008, Proceedings, Part IV 10}, pp.\  705--718. Springer, 2008.

\bibitem[Wei et~al.(2022)Wei, Tay, Bommasani, Raffel, Zoph, Borgeaud, Yogatama, Bosma, Zhou, Metzler, et~al.]{wei2022emergent}
Wei, J., Tay, Y., Bommasani, R., Raffel, C., Zoph, B., Borgeaud, S., Yogatama, D., Bosma, M., Zhou, D., Metzler, D., et~al.
\newblock Emergent abilities of large language models.
\newblock \emph{arXiv preprint arXiv:2206.07682}, 2022.

\bibitem[Zeiler \& Fergus(2014)Zeiler and Fergus]{zeiler2014visualizing}
Zeiler, M.~D. and Fergus, R.
\newblock Visualizing and understanding convolutional networks.
\newblock In \emph{Computer Vision--ECCV 2014: 13th European Conference, Zurich, Switzerland, September 6-12, 2014, Proceedings, Part I 13}, pp.\  818--833. Springer, 2014.

\bibitem[Zhang et~al.(2018)Zhang, Bargal, Lin, Brandt, Shen, and Sclaroff]{zhang2018top}
Zhang, J., Bargal, S.~A., Lin, Z., Brandt, J., Shen, X., and Sclaroff, S.
\newblock Top-down neural attention by excitation backprop.
\newblock \emph{International Journal of Computer Vision}, 126\penalty0 (10):\penalty0 1084--1102, 2018.

\bibitem[Zhang et~al.(2021)Zhang, Khakzar, Li, Farshad, Kim, and Navab]{zhang2021fine}
Zhang, Y., Khakzar, A., Li, Y., Farshad, A., Kim, S.~T., and Navab, N.
\newblock Fine-grained neural network explanation by identifying input features with predictive information.
\newblock \emph{Advances in Neural Information Processing Systems}, 34:\penalty0 20040--20051, 2021.

\bibitem[Zhou et~al.(2016)Zhou, Khosla, Lapedriza, Oliva, and Torralba]{zhou2016learning}
Zhou, B., Khosla, A., Lapedriza, A., Oliva, A., and Torralba, A.
\newblock Learning deep features for discriminative localization.
\newblock In \emph{CVPR}, pp.\  2921--2929, 2016.

\bibitem[Zhou et~al.(2022)Zhou, Booth, Ribeiro, and Shah]{zhou2022feature}
Zhou, Y., Booth, S., Ribeiro, M.~T., and Shah, J.
\newblock Do feature attribution methods correctly attribute features?
\newblock In \emph{Proceedings of the AAAI Conference on Artificial Intelligence}, volume~36, pp.\  9623--9633, 2022.

\end{thebibliography}
\bibliographystyle{icml2024}
\clearpage

\appendix


\begin{center}
    \LARGE \bf Appendix
\end{center}



\section{Experiment details for Figure~\ref{fig:trained_nn}}\label{sec:appendix:trained_nn_failure}

The experiment result is shown in Figure~\ref{fig:trained_nn}. In this experiment, we design a synthetic dataset that comprises two classes. In each class, we only construct one image as the only data point. Each image has two visual components in the image. The class $0$ image has one rectangular object in the center and another smaller rectangular object at the top left edge, while the class $1$ image has one round object in the center and another round object at the lower right edge. We consider two components as the ground truth of the task. After training on the data, the model can achieve $100\%$ accuracy. Then, we split every training image into two separate images, where one image contains the center object, and the other image contains the object at the edge. The split results are Test Data $1$ and Test Data $2$ in Figure~\ref{fig:trained_nn}. By using Test Data $1$ and Test Data $2$ as test images for the trained model, we observe that the model still achieves $100\%$ accuracy for Test Data $2$, but has $50\%$ accuracy (which is random guess in $2$-class case) for Test Data 1. Hence, we confirm that the trained model only learned to rely on edge objects for decision-making and completely ignored the centered object.

\section{Extended related work}\label{sec:appendix:related_work}

\subsection{Feature attribution}
This section provides a taxonomy of several feature attribution methods.

\subsubsection{Gradient-based}
The early methods relied on a linear approximation of the neural network around the input point. \cite{simonyan2013deep} assumed the model to be linear (inspired by the piecewise linearity of ReLU) and proposed using the gradient of output with respect to input as a saliency map. This work coined the term ``saliency map'' for attribution, and many following works used the same nomenclature. The gradient of neural networks with respect to images looked ``noisy''. Therefore, there were attempts to remove the "noise" from the images. One of these methods averages the gradient over a neighborhood around the input to smooth out the gradient. The SmoothGrad \cite{smilkov2017smoothgrad} method adds Gaussian noise to the image and generates multiple samples (in a Gaussian neighborhood of the input). Then computes the gradient and averages it. Other works proposed backpropagating only positive values to remove the "noise". Deconvolution \cite{zeiler2014visualizing} and subsequently Guided Backpropagation \cite{springenberg2015striving} and Excitation BackProp \cite{zhang2018top} work based on backpropagating positive values.

\subsubsection{Latent features}
The most famous work from this category is CAM/GradCAM \cite{selvaraju2017gradcam,zhou2016learning}. The method does a weighted average of activation values of the final convolutional layer. The weights are selected based on their gradient values. Another work \cite{khakzar2021neural} computes the input attribution based on the contribution of neurons to the output. This is achieved by computing the gradient of critical pathways. Another method restricts the flow of information in the latent space \cite{schulz2020restricting} and proposes information bottleneck attribution (IBA).

\subsubsection{Backpropagation of relevance}
The most important works within this category are LRP \cite{bach2015pixel}, and DeepLift \cite{shrikumar2017learning}. The LRP method provides a general framework for backpropagating the relevance to input while satisfying a conservation property (similar to Kirchoff's law of electricity). The relevance entering a neuron equals the relevance coming out. There are several rules of backpropagation within the LRP framework, each with its own properties. DeepLift proposes a chain rule for propagating output differences to the input. An improved version of DeepLift is the DeepSHAP \cite{lundberg2017unified} method, which we use in the experiments.

\subsubsection{Perturbation-based methods}
Another category of methods performs attribution by perturbing the input. The most straightforward way is to occlude/mask pixels (or patches of pixels) within the image and measure the output difference \cite{zeiler2014visualizing,ancona2018towards}. The output difference reflects the importance of the removed feature. It is also possible to search for a mask on the input image. For instance, searching for the smallest region within the image which preserves the output \cite{fong2019understanding}. 

\subsubsection{Shapley value}
The Shapley value \cite{shapley1953value} itself is based on the occlusion of features. It is a notion from cooperative game theory for assigning the contribution of players to a game. We can consider the pixels within the image as players and the output of the network as the score of the game. The Shapley value is a unique solution that satisfies the symmetry, dummy, linearity, and completeness axioms altogether. It is of exponential complexity and, therefore, almost impossible to compute for large images and neural networks. \cite{lundberg2017unified} first proposed using the Shapley value (an approximation) for feature attribution in neural networks by introducing the SHAP and DeepSHAP (for neural networks) methods. Another method based on the Shapley value is the integrated gradients \cite{sundararajan2017axiomatic}, which is the Aumann-Shapley value in the continuous domain \cite{ManyShapleySundararajan2020}.

\subsection{Feature attribution evaluation}
There have been many efforts to evaluate feature attribution methods. Each of these methods evaluates the attribution through a different lens and reveals unique insights.

\subsubsection{Alignment with human intuition}
Early feature attribution works \cite{simonyan2013deep,zeiler2014visualizing,zhang2018top,bach2015pixel} evaluated different saliency methods based on how they are aligned with what we humans think is salient. For instance, if a saliency method highlights a certain object, we might conclude that the network uses this object. However, the network may be using different features than ours. In this case, we wrongly consider the attribution method as correct (known as confirmation bias). Or in another case, the network might be using the object, but the saliency method might highlight something else. How would we know that the saliency is wrong? Several methods, such as the pointing game \cite{zhang2018top}, made this visual evaluation systematic by comparing the saliency values with ground truth bounding box annotations. Even though human evaluation is flawed (e.g., due to confirmation bias), it is still useful for debugging and better understanding attributions.

\subsubsection{Sanity checks}
Another group of metrics evaluates the sanity of methods. They check if the methods possess specific properties that attribution methods must have. \cite{adebayo2018sanity} checks what happens to attributions when we replace the network weights with random values. It is surprisingly observed that some methods generate the same saliency values when the network is randomized. Another work \cite{sixt2020explanations} checks how the attribution method changes when it is applied to a different output. In the presence of multiple classes within the image, the attribution method is expected to point to the relevant features of the explained class.

\subsubsection{Evaluation by perturbation}
The intuition behind these methods is aligned with the perturbation-based feature attribution methods. If a feature is contributing to the output, its removal should affect the output. And the more the effect, the more important the feature. Based on this intuition, \cite{samek2016evaluating} proposed perturbing/removing pixels based on their attribution score and plotting their effect as we keep removing them. In contrast to perturbing pixels according to a specific order, Sensitivity-N~\citep{ancona2018towards} randomly selects $N$ pixels to perturb, after that measuring the correlation between the change in prediction and the sum of attribution of the perturbed pixels. However, one issue with this approach is that the output score might be due to out-of-distribution values and not the relevance of the feature. Another method \cite{hooker2019benchmark} tries to remedy this issue by training the neural network from scratch on the perturbed dataset. In other words, a certain percentage of each image within the dataset is perturbed based on attribution scores, and the network is trained on this new dataset. The method then computes the accuracies in different perturbation percentages.

\subsubsection{Axiomatic}
Several axioms provide a framework to formalize the contribution of a feature \cite{ManyShapleySundararajan2020,lundberg2017unified,sundararajan2017axiomatic}. For instance, if the perturbation of a feature does not affect the output in any combination of features, then it is said to be a dummy feature. The dummy axiom demands that an attribution method should assign zero relevance to a dummy feature. Several other axioms reflect properties that we want the attributions to conform to. A combination of four properties, dummy, symmetry, linearity, and completeness, is only satisfied by the Shapley value \cite{shapley1953value, sundararajan2017axiomatic}. However, the Shapely value is not computable for large images and networks. Moreover, the proper reference value for removal must be chosen for computing the Shapley value. The complexity and the choice of reference values remain open problems. Moreover, for the existing methods, it is complex to theoretically show whether they conform to axioms as the methods are complex to analyze on neural networks. It is also shown that several axiomatic methods break the axioms in practical applications \cite{ManyShapleySundararajan2020}.

\subsubsection{Alignment with synthetic ground-truth}
An intriguing and promising solution is proposing synthetic settings to evaluate explanations. Several approaches propose generating datasets with known correlations between features and labels \cite{arras2022clevr,agarwal2023evaluating,zhou2022feature}. However, there is no guarantee that the network will use the same features to predict the outputs. 
To address this problem, \citet{khakzar2022explanations} proposes a post-hoc solution of inverse feature generation by generating input features using the network. However, this direction faces challenges related to generated out-of-distribution input features. It is still unclear which parts of the generated features contribute and how they contribute. Therefore, we will not have a ground truth of which feature is relevant to the network. This category is the focus of this work. For further information, please refer to the main text.

\section{Implementation details of synthetic models}\label{sec:appendix:detail_model_design}
In this section, we explain neural modules used in our synthetic model in finer detail. 
\subsection{Neural number checker}\label{sec:appendix:number_check}
\begin{figure*}[h]
    \centering
    \includegraphics[width=\textwidth]{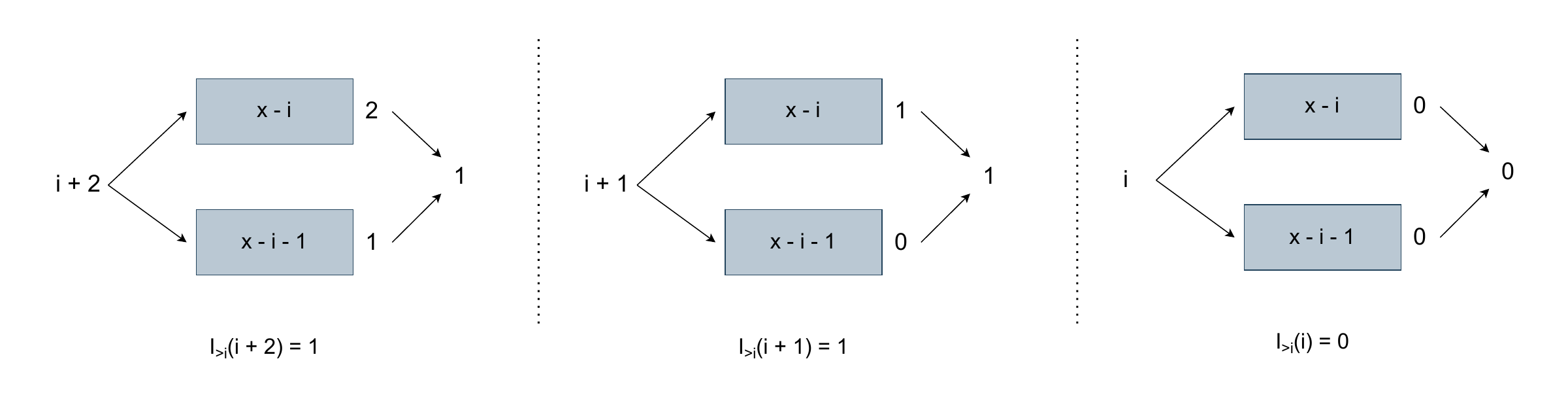}
    \caption{Example computational graph of $I_{>i}(x)$. Blue boxes symbolize neurons, with their respective computations indicated within the box. ReLU activation is applied after each neuron, which is omitted in the figure. We can set parameters depending on the $i$ value. We show three examples in the figure to show that this neural network structure activates by returning $1$ when the input is greater than $i$, otherwise, it outputs $0$.}
    \label{fig:appendix:larger_than_module}
\end{figure*}

We show the design of a neural number checker, which is used in other synthetic neural networks. The number checker is designed to take integer numbers as input. Furthermore, this module should only activate on one predefined number by returning $1$, otherwise, it should always return $0$. To achieve number checking in neural networks, we first design a network that examines whether a number is greater than a predefined number. For this task, we design 
\begin{equation}
    I_{>i}(x) = \relu(\relu(x-i) - \relu(x-i-1)).
\end{equation}
Figure~\ref{fig:appendix:larger_than_module} shows an example of this network. $I_{>i}(x)$ returns $1$ if x is greater than $i$, and returns $0$ otherwise.

\begin{figure*}[h]
    \centering
    \includegraphics[width=\textwidth]{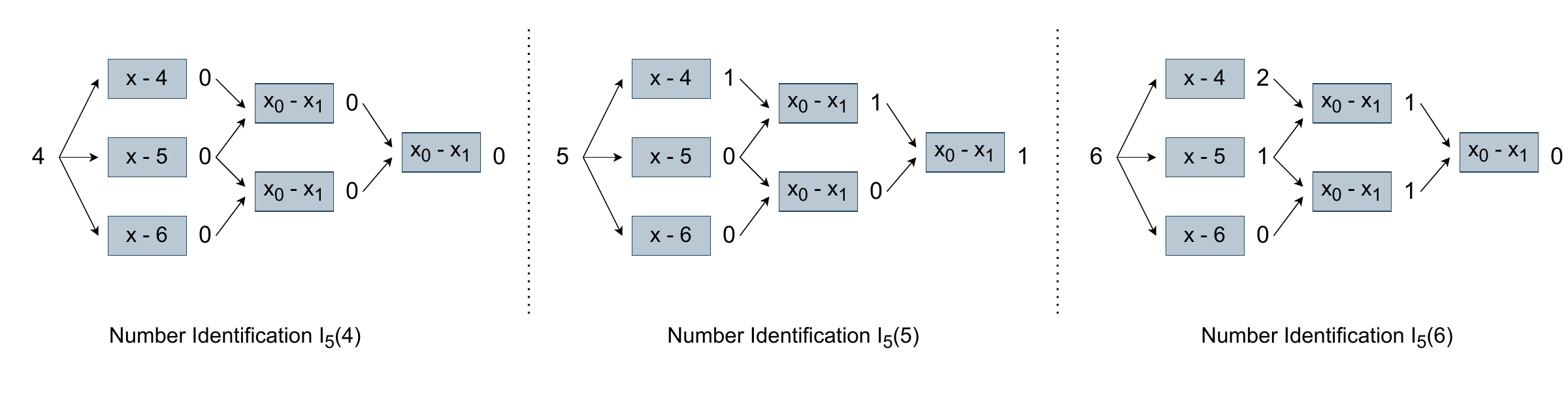}
    \caption{Example computational graph of $I_5(5)$. Blue boxes symbolize neurons, with their respective computations indicated within the box. ReLU activation is applied after each neuron, which is omitted in the figure. We show three cases where the input is $4$, $5$, and $6$. For natural numbers as inputs, this network only returns $1$ when the input is exactly $5$. We can alter parameters to enable the identification of any natural numbers. }
    \label{fig:appendix:check_number_module}
\end{figure*}

With the help of $I_{>i}(x)$, we can further construct a neural network that performs number identification. For this purpose, we constrain the input to this network to be integers. Then, number identification can be achieved by performing
\begin{equation}
    I_N(x) = \relu(I_{>N-1}(x) - I_{>N}(x))).
\end{equation}
The logic behind this design is that if $x$ is larger than both $N-1$ and $N$, then the output is $0$. However, if $x$ is larger than $N-1$, but not larger than N $N$, $I_N(x)$ returns $1$. For natural number inputs, only $N$ suffice the condition for outputting $1$. Figure~\ref{fig:appendix:check_number_module} shows an example network $I_5(x)$.

\subsection{CNN accumulator}\label{sec:appendix:accumulator}
In this section, we discuss how to construct CNNs that can perform pixel number accumulation. We design two types of CNNs for the accumulation task. One type of CNNs is initialized with uniform weights, while the other type can have non-uniform weights. We design these two types to show that our design can have both simple and complex CNN structures.

\begin{figure}[h]
    \centering
    \includegraphics[width=\columnwidth]{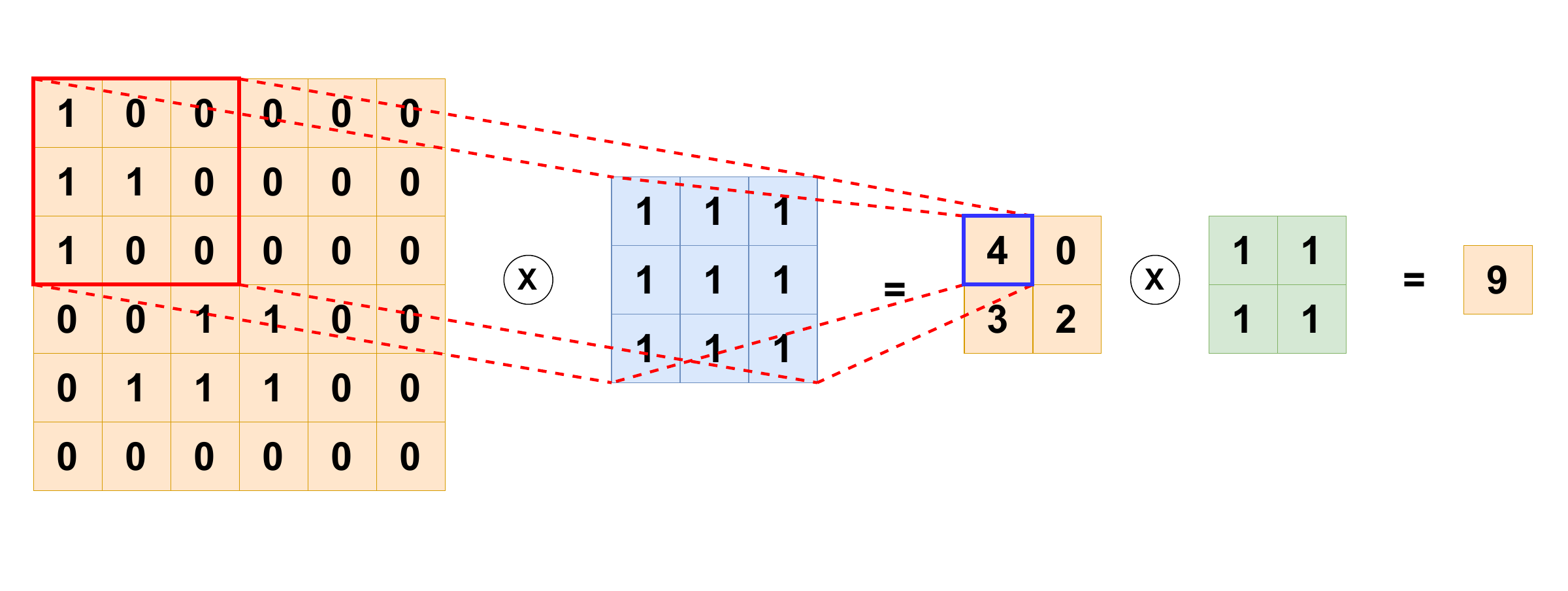}
    \caption{Example computation for an uniform CNN accumulator. ReLU activation is applied after each neuron, which is omitted in the figure. Given a single-channel input of the size $6 \times 6$, we can subsequently apply a uniform $3 \times 3$ CNN kernel and a $2 \times 2$ CNN kernel. The $1 \times 1$ output is the sum of input values. This example only has two CNN layers. For larger input sizes, we can design more CNN layers and kernels to perform accumulation. }
\label{fig:appendix:uniform_cnn_accumulation}
\end{figure}

\emph{Uniform weight initialization: }Realizing a CNN for pixel accumulation is intuitively simple. For a single-channel input, one can have a CNN module with multiple layers. Where each layer has only one kernel that sums up some neighboring pixels of the input. To realize this, the CNN kernel can have uniform weights of $1$, and have a stride size equal to its kernel size. By applying multiple such CNN layers, we can sum up all pixels in the input. Figure~\ref{fig:appendix:uniform_cnn_accumulation} shows an example of a uniform CNN layer for accumulating pixels.

\begin{figure}[h]
    \centering
    \includegraphics[width=\columnwidth]{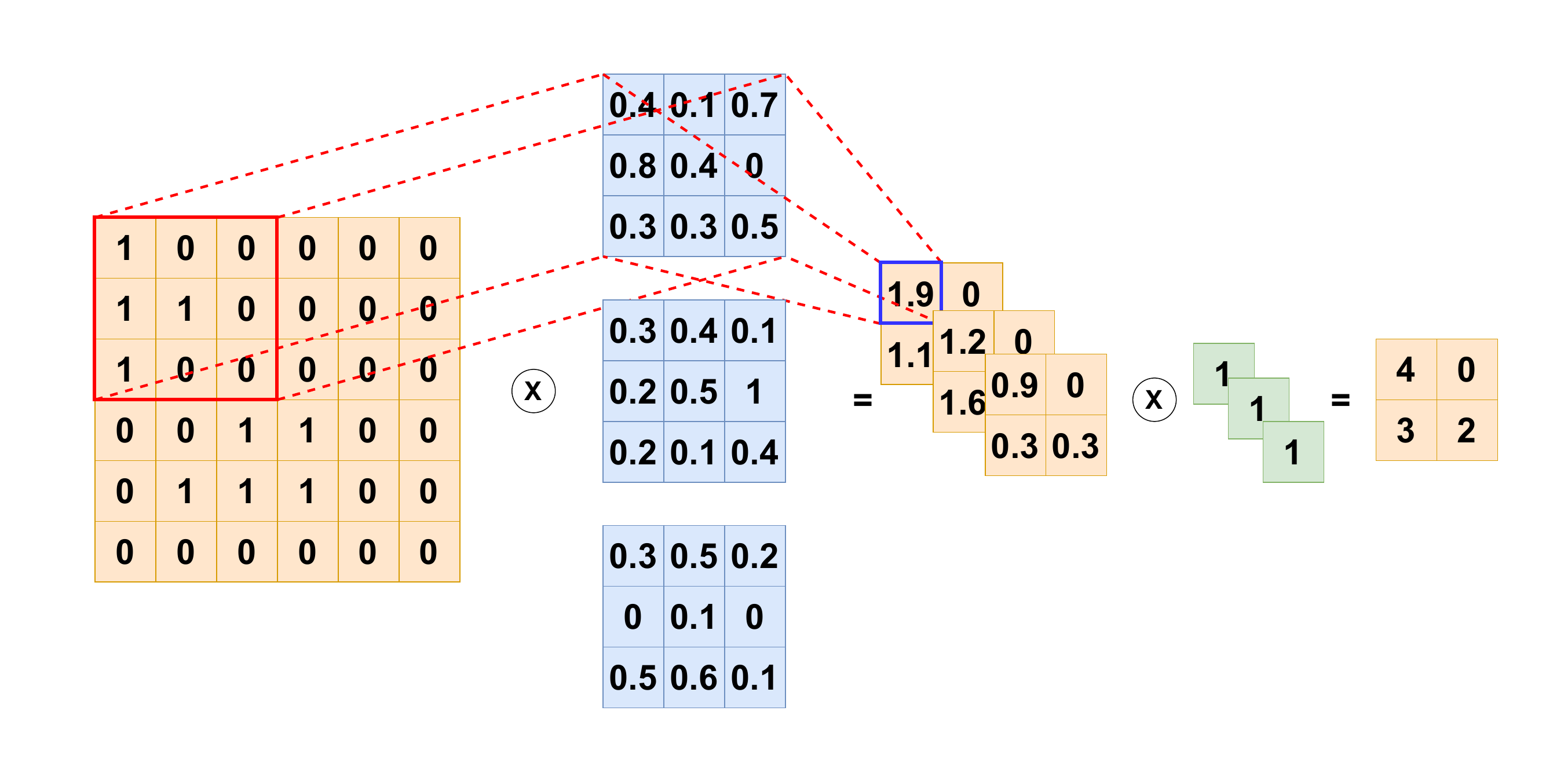}
    \caption{Example computation for a non-uniform CNN accumulation block. ReLU activation is applied after each neuron, which is omitted in the figure. The block consists of two CNN layers. In this example, the first layer has $3$ $3 \times 3$ CNN kernels, the second layer has one $1 \times 1$ CNN kernel. One accumulation block can sum up neighboring pixels within a fixed range without having uniform weights. }
\label{fig:appendix:non_uniform_cnn_accumulation}
\end{figure}

\emph{Non-uniform weight initialization: }The above example shows the construction of a CNN accumulator using uniform weights. However, due to random initialization and randomness in training, it is unlikely that a trained model has uniform weights. Hence, we realize another version of CNN accumulator that uses non-uniform weights. For a non-uniform CNN accumulator, we cannot use a single CNN layer to perform the number accumulation task, as only a uniform CNN layer can carry this task. Instead, we develop a CNN accumulation block using two CNN layers. The first CNN layer in this block has $N$ kernels with kernel weight $w_{ij}$ for the $j^{th}$ index of the $i^{th}$ kernel. Thus, after applying this layer, we will have $N$ feature maps. The second CNN layer has a $1\times 1$ CNN kernel, where its weight is defined as $m_i$ for the $i^{th}$ index. If the weight of these two layers satisfies $\forall j, \sum_i w_{ij}\cdot m_i = 1$, then the output still performs accumulation. Hence, we can initialize the accumulation block in a pseudo-random way, where we randomly set some of the weights, then calculate other weights that enable the CNN block to conduct accumulation. An example of this block is shown in Figure~\ref{fig:appendix:non_uniform_cnn_accumulation}. In Figure~\ref{fig:appendix:non_uniform_cnn_accumulation}, we simplify the design by using uniform weights for the second CNN layer. By subsequently connecting accumulation blocks, we can perform the accumulation task with a CNN module that has non-uniform kernel weights.

\subsection{Neural modulo module}\label{sec:appendix:modulo}
\begin{figure}[h]
    \centering
    \includegraphics[width=\columnwidth]{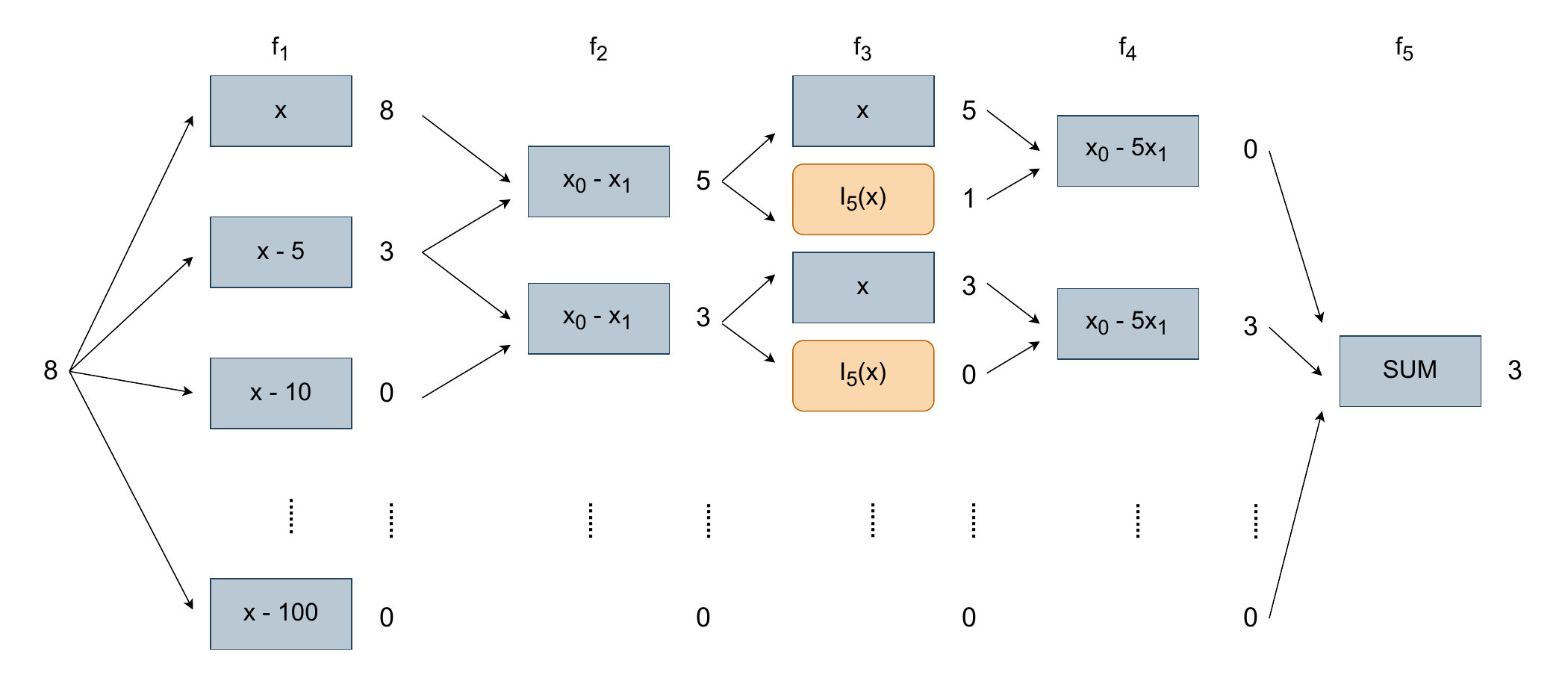}
    \caption{Example computational graph of $f_{\text{modulo}, 5}(8)$. Blue boxes symbolize neurons, with their respective computations indicated within the box. ReLU activation is applied after each neuron, which is omitted in the figure.}
    \label{fig:appendix:modulo_module}
\end{figure}

In this section, we provide an in-depth explanation of the designed model for modulo calculation. We explain the functionality of each MLP layer.
The first layer of the module is 
\begin{equation}
    f_1(x) = \relu(x, x-N, x-2N, ..., x-\lceil\frac{U}{N}\rceil N),
\end{equation}
where $U$ represents the maximal input value that the module can process, which can be scaled according to available computation resources. The incentive of applying $f_1(x)$ is to get $x \mod N$, $N + x \mod N$, $2N + x \mod N$, ..., until $x$. The calculation of $f_1(x)$ is valid for any natural number input that is not larger than $N$. The second layer is defined as 
\begin{equation}
    f_2(\vec{x}) = \relu(x_0 - x_1, x_1 - x_2, ...).
\end{equation}
Function $f_2(\vec{x})$ performs differentiation between subsequent outputs of $f_1(x)$. For all non-zero outputs of $f_1(x)$, their difference is always $N$. Furthermore, the smallest non-zero output is always $x\mod N$. Thus, we would have $f_2(f_1(x)) = (N, N, ..., x  \mod  N, 0, ..., 0)$ after propagating through the first two layers. 
At this stage, our output only comprises three possible numbers: $N$, $x \mod N$, and $0$. Next, we eliminate the number $N$ in the output vector. This is accomplished by first checking if an output is $N$, then eliminating $N$, which is done by applying 
\begin{equation}
    f_3(\vec{x}) = \relu(x_0, I_N(x_0), x_1, I_N(x_1), ...), 
\end{equation}
and 
\begin{equation}
    f_4(\vec{x}) = \relu(x_0 - N x_1, x_2 - N x_3, ...).
\end{equation}
$f_3(\Vec{x})$ is responsible for checking all outputs after $f_2(\Vec{x})$, if an output is $N$, it appends a flag number $1$ after this output. Otherwise, if an output is not $N$, which is either $x \mod N$ or $0$, it appends a flag number $0$. $f_4(\vec{x})$ Subtracts each output of $f_2(\vec{x})$ with $N$ times its flag number obtained from $f_3(\vec{x})$. Hence, we have $f_4(f_3(f_2(f_1(x)))) = (0, 0, ..., x  \mod  N, 0, ...)$. 
Lastly, we have one layer 
\begin{equation}
f_5(\vec{x}) = \relu(\sum_{i=0}x_i),
\end{equation}
to accumulate the output vector. Therefore, as shown in Figure~\ref{fig:appendix:modulo_module}, we can obtain the modulo calculation with 
\begin{equation}
    f_{\text{modulo}, N}(x) = f_5(f_4(f_3(f_2(f_1(x))))) = x\mod N. 
\end{equation}

\subsection{Neural color detector module}\label{sec:appendix:color_detector}
This section describes how we design a CNN part that can detect certain colors. Since we want to check every pixel in an image, we design CNN modules with $1 \times 1$ kernels for color detection. For this module, we assume the input image is in RGB format, and the output of this module has dimensions of $N\times H \times W$, where $N$ denotes the number of colors we want to detect, $H$ and $W$ are the height and the width of the input image. Each of the $N$ output channels is responsible for the detection of a predefined target color. Specifically, if a pixel in the input corresponds to the $i^{th}$ target color the model aims to detect, then the neuron at the same position in the $i^{th}$ channel of the output should be 1. Otherwise, the corresponding output neuron should be 0.
\begin{figure*}[h]
    \centering
    \includegraphics[width=\textwidth]{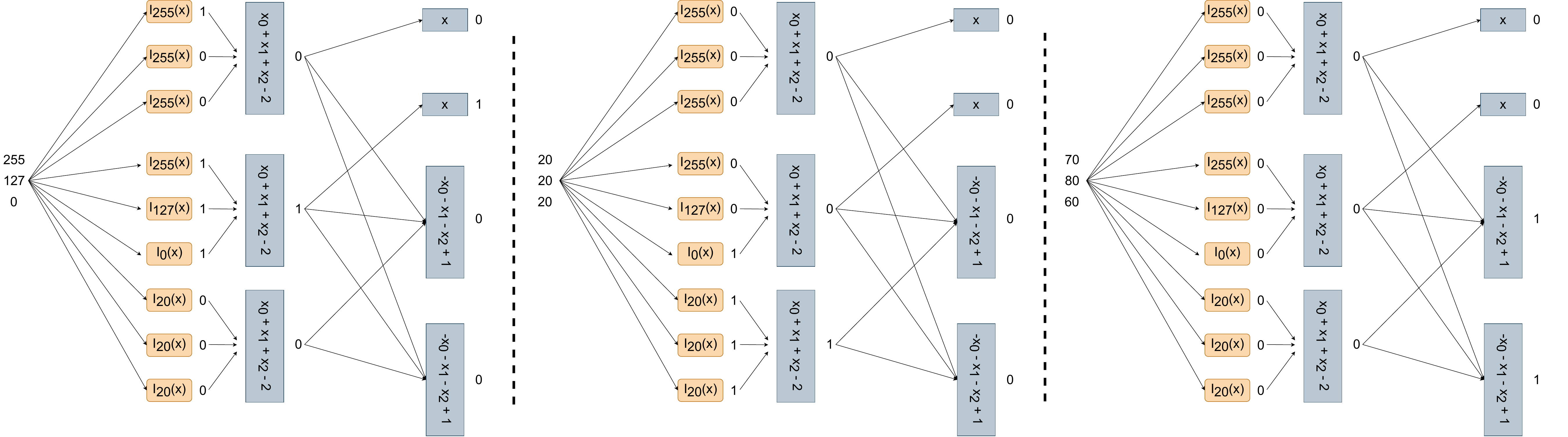}
    \caption{Example computational graph for color detection. Blue boxes symbolize neurons, with their respective computations indicated within the box. ReLU activation is applied after each neuron, which is omitted in the figure. This example shows a color detector with two redundant channels. The color detector is responsible for detecting target colors $(255, 255, 255)$ and $(255, 127, 0)$. Moreover, the color detector is programmed to ignore the background color, which is $(20, 20, 20)$. In the left case, one of the target colors is detected. Hence, the detector activates the target color channel. In the middle case, none of the target colors is detected. However, the input is the same as the predefined background color $(20, 20, 20)$. Subsequently, the input is considered in distribution, and no output channel is activated. In the right case, the input is not one of the target colors or the background color, so all redundant channels are activated in response to OOD inputs. We show the computation in fully connected NN layers. Nevertheless, color detectors are realized as CNN layers with $1\times 1$ kernels in our implementation.}
    \label{fig:appendix:color_detector}
\end{figure*}
To accomplish this, the module should be able to check all three input channels and compare if the RGB values match the color to be detected. As discussed in Section~\ref{sec:method:design}, we leverage 
\begin{equation}
    C(r, g, b) = \relu(I_R(r) + I_G(g) + I_B(b) - 2)
\end{equation}
for detecting RGB values of pixels. $C(r, g, b)$ only outputs $1$ if all three input values match the RGB value of the color to be detected. Illustration of $C(r, g, b)$ is shown in Figure~\ref{fig:comp_graph}. To enable simultaneous detection of multiple target colors, we concatenate several $C(r, g, b)$ with their weights set for different RGB values. This is shown as the first two layers of the network in Figure~\ref{fig:appendix:color_detector}.

We further elaborate on how to create redundant channels in our color detector design. Instead of outputting the activation map of the dimension $N \times H \times W$, the output can have the shape of $(N + R) \times H \times W$, where $R$ denotes the number of redundant channels. Redundant channels serve as OOD channels that only activates if OOD colors are present in an image. We define OOD colors to be any color other than our target colors and the predefined background color. To activate redundant channels, we use
\begin{equation}
    R(r, g, b) = \relu( - \sum C_i(r, g, b) + 1), 
\end{equation}
where $C_i(r, g, b)$ indicates color detection for different colors. If none of any $C_i(r, g, b)$ is activated, then $R(r, g, b)$ returns $1$. In the example shown in Figure~\ref{fig:appendix:color_detector}, the last two outputs are redundant channels. 


\section{Implementation of Single-color-modulo Setting}\label{sec:appendix:single-color-modulo-setting}

Given monochromatic images as input, we manually program a synthetic neural network to count the number of white pixels in the image (the Accumulator) and output the modulo $N$ of the total pixel count (the Modulo), where the divisor $N$ is a predefined number.
The design of these images consists of multiple pixel patches of arbitrary shape within a constant background (reference value). 
The Accumulator and Modulo programs have \emph{properties} tailored toward evaluating feature attributions:
In our model design, the addition/removal of any ground truth (white) pixel to/from the background (black) equally affects the count of white pixels. Hence, we know each and every white pixel is equally relevant.


To develop a neural network model that can perform this task, we design a model that consists of two components. The first component is a Convolutional Neural Network (CNN) tasked with counting the total number of white pixels in the input image. This CNN accumulation module can be initialized with either uniform or non-uniform weights. 
To initialize this CNN with uniform weights, we can simply set all weights of CNN kernels to $1$. We further elaborate on the initialization of non-uniform weights in Appendix~\ref{sec:appendix:accumulator}. The second part of the model is a multilayer perceptron (MLP) designed to perform the modulo operation. 
We formally define the modulo module using functions. Firstly, we leverage the number identification module $I_N(x)$ to further build up the neural modulo module. The first layer of the module is $f_1(x) = \relu(x, x-N, x-2N, ..., x-\lceil\frac{U}{N}\rceil N)$, where $U$ represents the maximal input value that the module can process, which can be scaled according to available computation resources. Note that $f_1$ produces a vector output. The second layer is defined as $f_2(\vec{x}) = \relu(x_0 - x_1, x_1 - x_2, ...)$. Thus, we would have $f_2(f_1(x)) = (N, N, ..., x  \mod  N, 0, ..., 0)$ after propagating through the first two layers. 
At this stage, our output only comprises three possible numbers: $N$, $x \mod N$, and $0$. Next, we eliminate the number $N$ in the output vector. This is accomplished by applying $f_3(\vec{x}) = \relu(x_0, I_N(x_0), x_1, I_N(x_1), ...)$, and $f_4(\vec{x}) = \relu(x_0 - N x_1, x_2 - N x_3, ...)$. Hence, we have $f_4(f_3(f_2(f_1(x)))) = (0, 0, ..., x  \mod  N, 0, ...)$. 
Lastly, we have one layer $f_5(\vec{x}) = \relu(\sum_{i=0}x_i)$ to accumulate the output vector. Therefore, as shown in Figure~\ref{fig:comp_graph}, we can obtain the modulo calculation with 
\begin{equation}
    f_{\text{modulo}, N}(x) = f_5(f_4(f_3(f_2(f_1(x))))) = x\mod N. 
\end{equation}
The complexity of the MLP model ensures that the synthetic model poses a moderate challenge for attribution methods. Further details regarding the dataset generation and model initialization can be found in Appendix~\ref{sec:appendix:number_check}, Appendix~\ref{sec:appendix:modulo}.

\section{Experiment detail for Table~\ref{table:model_summary}}\label{sec:app:model_summary_detail}
The model summary is conducted on two models. 

Ours is the designed model for the single-color-modulo setting. The model is designed to have two modules, one CNN model for pixel accumulation, and a Linear module for modulo calculation. 

The CNN module comprises 10 CNN layers. The linear module has of 7 linear layers as described in the Appendix~\ref{sec:appendix:modulo}. ResNet-8 model has 3 residual blocks with 2 CNN layers in each residual block, a CNN layer at the beginning, and a linear layer at the end.


\section{Experiments in the single-color-modulo setting}
\label{sec:appendix:exp_black_white_bezier}

\subsection{Synthetic dataset generation}\label{sec:appendix:exp_black_white_bezier_dataset_generation}
\begin{figure*}[t]
    \centering
    \includegraphics[width=\textwidth]{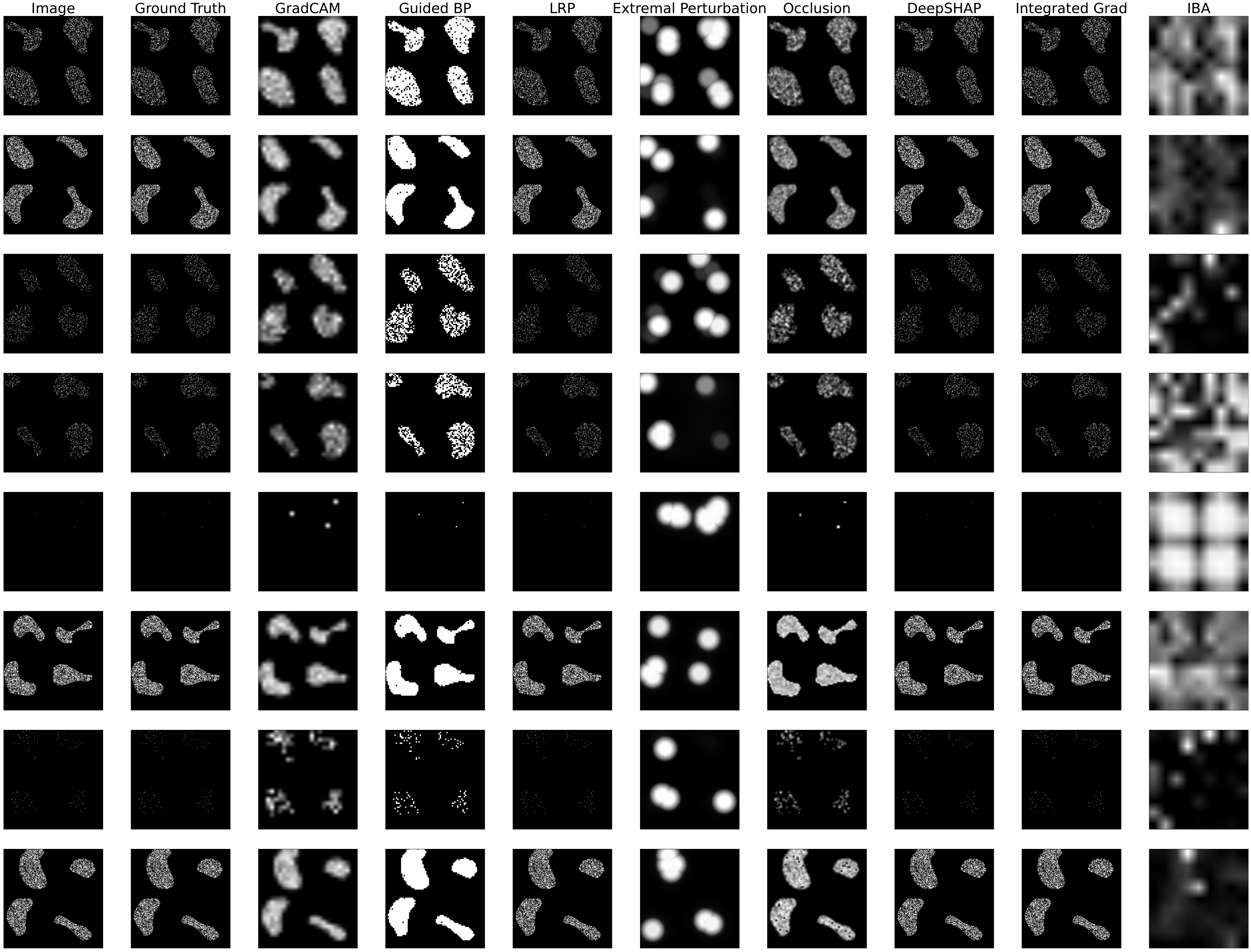}
    \caption{Visualization of images and attribution maps in single-color-modulo setting.}
    \label{fig:appendix:visual_result_black_white_bezier}
\end{figure*}

\textbf{Image generation:} We place four non-overlapping patches at four randomly selected centers within each image, with each patch surrounded by B\'ezier curves. Pixel intensities in the patches are randomly sampled from the Bernoulli distribution $Bernoulli(0.5)$. To save the generated images in single-channel PNG format, we scale the pixel value of $1$ to 255. Samples of generated images are shown in the first column of Figure~\ref{fig:appendix:visual_result_black_white_bezier}.

\textbf{Ground truth label:} We predefine a modulus $N$ (default value $30$). Let $s$ represent the sum of all $1$s in a synthetic image; the ground truth label for the image is then determined as $s \mod N$. For example, if an image contains $100$ pixels of $1$ and $N = 30$, the ground truth label is $10$.

\textbf{Ground truth features:} For white pixels in the image, pixels have a uniform ground truth feature importance of $1$, whereas for black pixels, they have a uniform ground truth feature importance of $0$.

We visualize some randomly selected samples in Figure~\ref{fig:appendix:visual_result_black_white_bezier}, including the generated images, ground truth features, and attribution maps.

\subsection{Hyper-parameters of attribution methods}

In this subsection, we present the key hyperparameters for various attribution methods to ensure reproducibility in future studies. The visual results of these attribution methods are presented in Figure~\ref{fig:appendix:visual_result_black_white_bezier}.

\begin{itemize}
    \item GradCAM: GradCAM is attached to the sixth layer of the CNN accumulator, specifically at \texttt{accumulator.layers.5} in our implementation.
    \item GuidedBP: Backpropagation is performed from the scalar modulo output.
    \item LRP: Similar to GuidedBP, backpropagation is performed from the scalar modulo output.
    \item ExPerturb: A perturbation area of 0.1 is used, with Gaussian blurring applied for image perturbation. The Gaussian blurring sigma is set to 21.0.
    \item Occlusion: A sliding window of shape $(1, 5, 5)$ and strides $(1, 3, 3)$ is employed, along with an all-zero baseline.
    \item DeepSHAP: An all-zero baseline is used.
    \item IG: An all-zero baseline is applied.
    \item $\text{IG}^{*}$: This method is equivalent to IG, as the ground truth baseline is 0.
    \item IBA: The information bottleneck is attached to the twelfth layer of the CNN accumulator, specifically at \texttt{accumulator.layers.11} in our implementation. The weight of the information loss is set to 20.
    \item Random: Random attribution values are sampled from a uniform distribution $\mathcal{U}(0, 1)$.
    \item Constant: The attribution values are set to a constant value of $1.0$.
\end{itemize}


\subsection{Ground-truth-based evaluation}
\label{sec:appendix:gt_eval_black_white_bezier_non_uniform}

\begin{figure*}[t]
    \centering
    \includegraphics[width=0.6\textwidth]{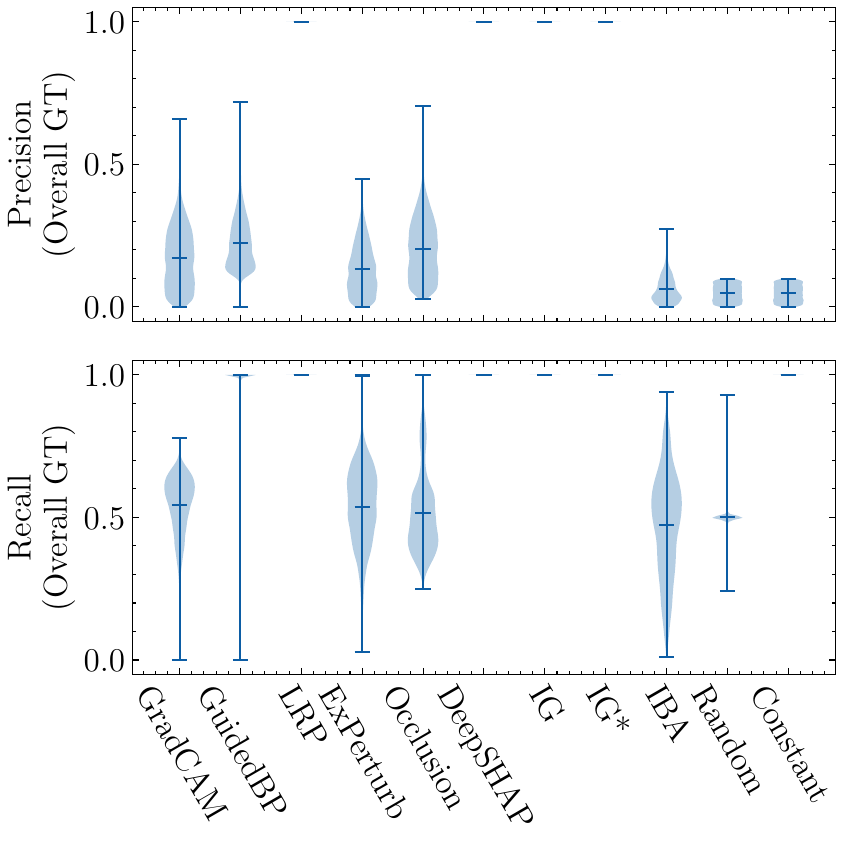}
    \caption{Ground-truth-based evaluation in the single-color-modulo setting. The maximal, minimal, and average values are marked with horizontal bars.}
    \label{fig:appendix:gt_eval_black_white_bezier_non_uniform}
\end{figure*}

\subsubsection{Experimental setup} 
We compute the \emph{overall} precision, recall based on the synthetic ground truth (feature importance) masks. Note that some attribution method can return attribution values substantially greater than $1.0$, which are not in the same scale as the ground truth value $1.0$. In this case, if we apply the formula of recall in Section~\ref{sec:method:transparent_env}, we would get a recall greater than $1.0$. To avoid this undesirable outcome, we normalize the attribution map as follows: 
\begin{equation}
    \label{eq:appendix:norm_attr_maps}
    \begin{aligned}
        t(a_j) = \begin{cases}
            \frac{a_j}{a_{+}} \quad & \text{ if } a_j \geq 0; \\
            \frac{a_j}{|a_{-}|} \quad & \text{ else},
        \end{cases}
    \end{aligned}
\end{equation}

where $a_j$ is the attribution associated with $j^{th}$ feature, $a_{+} = \max_j a_j$ for all $a_j \geq 0 $, and $a_{-} = \min_j a_j$ for all $a_j < 0$, respectively. The attribution value after normalization is within the interval $[-1, 1]$.

\subsubsection{Experimental results} Figure~\ref{fig:appendix:gt_eval_black_white_bezier_non_uniform} presents violin plots of overall precision and recall. The three horizontal bars within each plot correspond to maximum, average, and minimum values, respectively. These plots indicate that LRP, DeepSHAP, and IG (equivalent to $\text{IG}^{*}$ in this setting) achieve optimal precision and recall for nearly all data samples. Although GuidedBP obtains high recall for most samples, its average precision is suboptimal. This observation is supported by the visual results, where GuidedBP mistakenly identifies black pixels within the B\'ezier patches as contributing pixels. Furthermore, the low precision of methods such as GradCAM and IBA is often attributed to their process of performing attribution on a neural network's hidden layer and subsequently resizing and interpolating the attribution maps to match the input image's spatial dimensions. This additional post-processing step frequently leads to over-blurring of attribution maps.

\begin{table*}[t]
    \centering
    \caption{Rankings of attribution methods in the single-color-modulo setting when evaluated with ground truth masks, Insertion/Deletion, and Sensitivity-N. Correlation denotes the Spearman's rank correlation with the \Fone-score (computed using the overall ground truth masks).}
    \begin{tabular}{c c c c c}
        \toprule
        \multirow{2}{*}{Attribution methods} & \multicolumn{4}{c}{Ranking} \\
        \cmidrule{2-5}
         & \Fone-score & Insertion & Deletion & Sensitivity-N \\
        \midrule
        GradCAM & 6 & 6 & 6 & 6 \\
        GuidedBP & 4 & 5 & 5 & 4 \\
        LRP & 0 & 3 & 0 & 0 \\
        ExPerturb & 7 & 7 & 7 & 7 \\
        Occlusion & 5 & 4 & 4 & 5 \\
        DeepSHAP & 1 & 0 & 1 & 1 \\
        IG & 2 & 1 & 2 & 2 \\
        $\text{IG}^{*}$ & 3 & 2 & 3 & 3 \\
        IBA & 8 & 8 & 8 & 8 \\
        Random & 10 & 10 & 9 & 10 \\
        Constant & 9 & 9 & 10 & 9 \\
        \midrule
        Correlation & $-$ & 0.94 & 0.98 & 1.00 \\
        \bottomrule
    \end{tabular}
    \label{tab:appendix:ranking_table_w_ig_star_black_white_bezier}
\end{table*}


\subsection{Insertion/Deletion}
\label{sec:appendix:ins_del_black_white_bezier_non_uniform}

\subsubsection{Experimental setup} 
In the single-color-modulo setting, directly comparing modulo results between perturbed and original images is insufficient for determining true pixel contribution, as perturbing a specific number of pixels can produce the same modulo result. For instance, if the sum of $1$s in perturbed pixels is divisible by the modulus $N$, the change in the modulo result is 0, suggesting no contribution from the perturbed pixels. To address this, we adapt the implementations of these metrics without compromising their correctness. We perturb pixels individually and compare model outputs at consecutive steps. Additionally, we use zero-intensity pixels as replacement pixels during the progressive perturbation process. Different modulo results indicate genuine pixel contribution and the attribution method's accurate identification of the contributing pixel at that step. Otherwise, the step is deemed a failure for the attribution method. We accumulate correct steps as a substitute for prediction variation in the original Insertion/Deletion or Sensitivity-N. In fact, the accumulated correct steps equal the number of perturbed pixels with value $1$. To compute the AUC in Insertion/Deletion, we normalize the number of correct steps by dividing it by the total contributing pixels in the ground truth mask, yielding a range of $[0, 1]$.

\subsubsection{Experimental results} 
As illustrated in Table~\ref{tab:appendix:ins_del_black_white_modulo_non_uniform}, the best performing methods are IG (or $\text{IG}^{*}$), DeepSHAP, and LRP, achieving optimal Insertion AUC and Deletion AUC. As indicated by the Spearman's rank correlations in Tables~\ref{tab:main:spearmans_rank_correlation} and~\ref{tab:appendix:ranking_table_w_ig_star_black_white_bezier}, the Insertion/Deletion evaluation is highly consistent with the ground-truth-based evaluation. 
It is important to note that the optimal Insertion AUC in Table~\ref{tab:appendix:ins_del_black_white_modulo_non_uniform} is not $1.0$, and the optimal Deletion AUC is not $0.0$. This occurs because we remove one pixel at each step, and the best Insertion curve is obtained when all $1$s in the image receive higher attribution than all $0$s. In this case, the Insertion curve is a monotone increasing straight line followed by a saturated flat line at $1.0$. Similarly, the optimal Deletion curve occurs when all the $1$s are removed before all the $0$s, resulting in a monotone decreasing straight line followed by a stagnant flat line at $0.0$. Consequently, the AUC of the optimal Insertion curve is smaller than $1.0$, and the AUC of the optimal Deletion curve is greater than $0.0$. The optimal AUCs depend on the number of pixels with value $1$ and the total number of pixels.

\begin{table*}[t]
    \centering
    \caption{Insertion/Deletion evaluation results in the single-color-modulo setting. A higher Insertion AUC and lower Deletion AUC indicate a better attribution method.}
    \begin{tabular}{c c c}
        \toprule
        Attribution Method & Insertion AUC $\uparrow$ & Deletion AUC $\downarrow$ \\
        \midrule
        GradCAM & 0.8827 & 0.1174 \\
        GuidedBP & 0.9015 & 0.0983 \\
        LRP & 0.9754 & 0.0246 \\
        ExPerturb & 0.7845 & 0.2157 \\
        Occlusion & 0.9151 & 0.0849 \\
        DeepSHAP & 0.9754 & 0.0246 \\
        IG & 0.9754 & 0.0246 \\
        IG* & 0.9754 & 0.0246 \\
        IBA & 0.6458 & 0.3526 \\
        Random & 0.4996 & 0.5009 \\ 
        Constant & 0.4447 & 0.4439 \\
        \bottomrule
    \end{tabular}    \label{tab:appendix:ins_del_black_white_modulo_non_uniform}
\end{table*}


\subsection{Sensitivity-N}
\label{sec:appendix:sens_n_black_white_bezier_non_uniform}

\begin{figure}[t]
    \centering
    \includegraphics[width=\columnwidth]{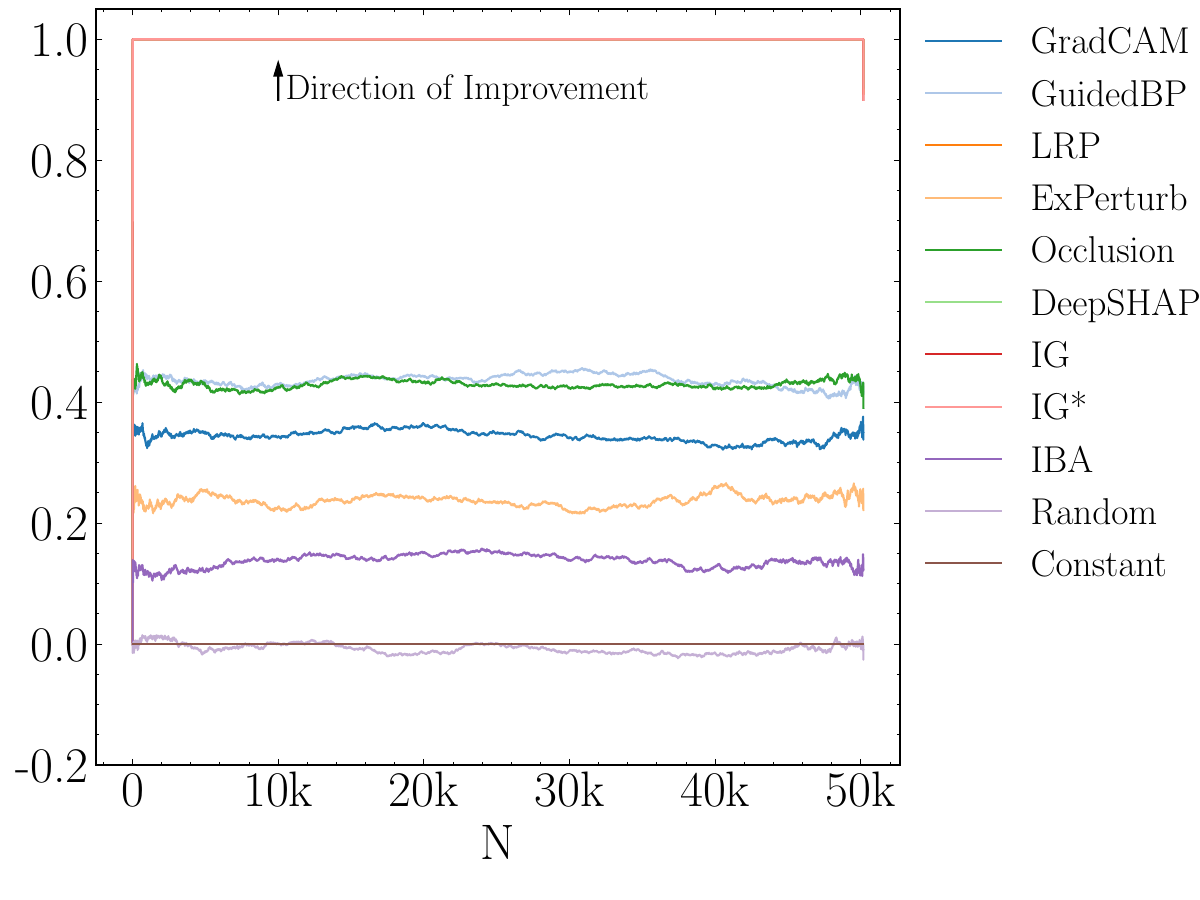}    
    \caption{Sensitivity-N evaluation results in the single-color-modulo setting. $y$-axis represents the Pearson Correlation Coefficient between the model output change and the total attribution of perturbed pixels.} \label{fig:appendix:sens_n_black_white_modulo_non_uniform}
\end{figure}

\subsubsection{Experimental setup} 
In standard Sensitivity-N, we randomly select $N$ pixels at each step. However, this approach is not applicable in the single-color-setting due to the same reason we explained previously in Insertion/Deletion. Therefore, following our dapted Insertion/Deletion in the single-color-modulo setting, we measure the model output change in an accumulative manner. Specifically, we need to compare the modulo results of two consecutive steps. Note that we can determine whether the $i^{th}$ pixel is contributing only when it is the sole pixel in which the perturbed pixels at two consecutive steps differ. In other words, we need to randomly select an \emph{additional} pixel to perturb compared to the previously perturbed pixels. Then, we record the model output change and the total attribution of perturbed pixels, respectively. We repeat the random selection for $100$ times and concatenate the recorded output change and total attribution into two vectors, respectively. Next, the Pearson Correlation Coefficient is computed between these two vectors. Subsequently, we increase the number of perturbed pixels $N$ from $1$ to the image size $224 \times 224$ and repeat the above procedure.

\subsubsection{Experimental results}
For readability, we reproduce the Sensitivity-N figure from Section~\ref{sec:exp:eval_of_eval_metrics} in Figure~\ref{fig:appendix:sens_n_black_white_modulo_non_uniform}. DeepSHAP, IG (or $\text{IG}^{*}$), and LRP achieve (near) optimal performance. This is consistent with the Insertion/Deletion results in Table~\ref{tab:appendix:ins_del_black_white_modulo_non_uniform} and ground-truth-based evaluation results in Figure~\ref{fig:appendix:gt_eval_black_white_bezier_non_uniform}.


\section{Experiments in the multi-color-sum setting}
\label{sec:appendix:exp_multi_color_sum_non_uniform}
In this section, we present more experimental results and explanations as a supplement to Section~\ref{sec:exp:eval_of_attr_methods} and Section~\ref{sec:exp:eval_of_eval_metrics}. Specifically, we present additional experimental results on the Multi-color-sum setting.

\subsection{Synthetic dataset generation}
\label{sec:appendix:data_multi_color_sum_non_uniform}
\begin{figure*}[t]
    \centering
    \includegraphics[width=\textwidth]{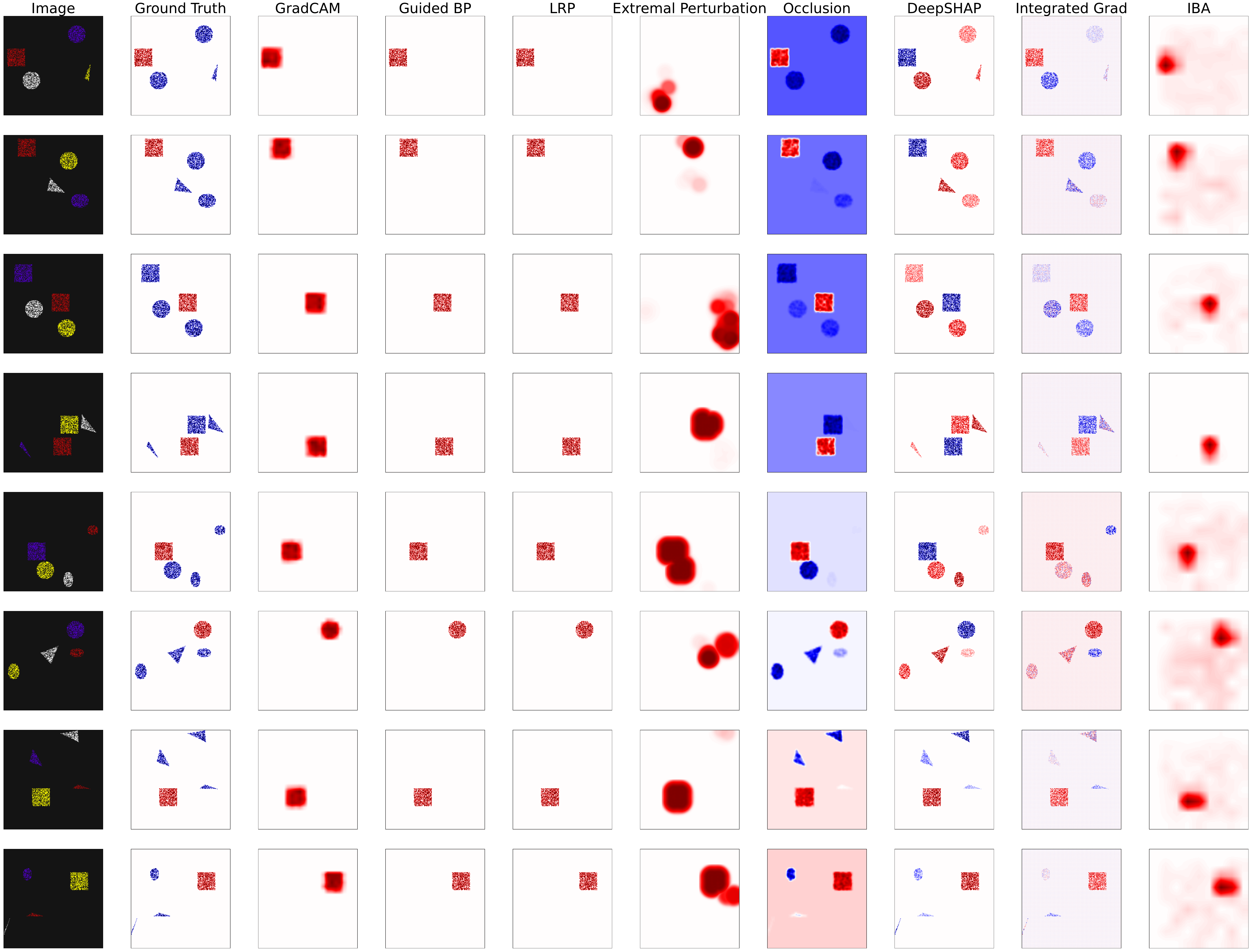}
    \caption{Visualization of images and attribution maps in multi-color-sum setting.}
    \label{fig:appendix:visual_result_multi_color_sum}
\end{figure*}

\textbf{Image generation: }This setting is designed to simulate a multi-class classification task. We randomly place four non-overlapping patches on an empty background image with a uniform intensity of $(20, 20, 20)$ (unsigned 8-bit \texttt{int}) in RGB format. Patch shapes are sampled from three categories: triangle, square, or circle. Additionally, patch sizes are randomly sampled. For each pixel within each patch, we independently and randomly sample a random variable from the $Bernoulli(0.5)$ distribution. If the sampled value is $1$, the pixel is set to the background color; otherwise, it is set to a unique color associated with the corresponding patch. Ultimately, an image will contain four distinct foreground colors. Some randomly selected samples are visualized in Figure~\ref{fig:appendix:visual_result_multi_color_sum}.

\textbf{Ground truth label:} The task is a multi-class classification task, where each class represents a non-background color. For each randomly generated image, the winning color in the image is the one with the most pixels of that color, and the image is labeled by the class of the winning color.

\textbf{Ground truth features:} The greater the number of pixels a color has, the more likely it is to become the winning color. Moreover, each pixel contributes equally when being counted. The corresponding output of a color increases by $1$ if we insert a pixel of that color into the image. For a non-winning color, increasing its number of pixels raises the likelihood of it becoming the winning color, thereby decreasing the probability of the current winning color. Background pixels do not contribute to the current winning color, as they are not detected by the color detector. Based on these principles, we establish the ground truth feature contribution masks as follows: (1) all pixels belonging to the winning color have a uniform contribution of $1$; (2) all pixels belonging to the non-winning colors have a uniform contribution of $-1$; and (3) all background pixels have a uniform contribution of $0$. As shown in Figure~\ref{fig:main:visualize_multi_color_sum}, positively contributing pixels are displayed in red, negatively contributing pixels are displayed in blue, and non-contributing pixels are displayed in white.

\subsection{Hyper-parameters of attribution methods}

In this subsection, we present the key hyper-parameters of attribution methods employed in this setting:

\begin{itemize}
    \item GradCAM: GradCAM is attached to the sixth layer of the CNN accumulator (specifically, \texttt{accumulator.layers.5} in the implementation).
    \item GuidedBP: Backpropagation is computed from the output after the softmax layer.
    \item LIME: The default segmentation method is Quickshift~\cite{vedaldi2008quick}.
    \item ExPerturb: We select a perturbation area of $0.1$ and employ Gaussian blurring for perturbation.
    \item Occlusion: A sliding window of shape $(3, 5 ,5)$ and strides $(3, 3, 3)$ is used, along with the ground truth baseline. Furthermore, the output change is measured at the layer before the softmax, as suggested by the original paper~\cite{zeiler2014visualizing}.
    \item DeepSHAP: The output change is measured after the softmax layer, and an all-zero baseline is employed.
    \item IG: The output change is measured after the softmax layer, and the ground truth baseline is employed.
    \item IBA: The information bottleneck is inserted into the eighth layer of the CNN accumulator (specifically, \texttt{accumulator.layers.7} in the implementation). Additionally, the weight of information loss is set to $20$.
\end{itemize}


\subsection{Ground-truth-based evaluation}
\label{sec:appendix:gt_eval_multi_color_sum_non_uniform}
\subsubsection{Experimental setup} 
We normalize the attribution maps in the same way as in Equation~\ref{eq:appendix:norm_attr_maps}. Then, the precision, recall, and \Fone-score are computed based on three sets of the designed ground truth masks, denoted with overall, positive, and negative ground truth (GT), respectively. Note that not all attribution methods can return negative attribution.

\subsubsection{Experimental results} 
\begin{figure*}[t]
    \centering
    \includegraphics[width=\textwidth]{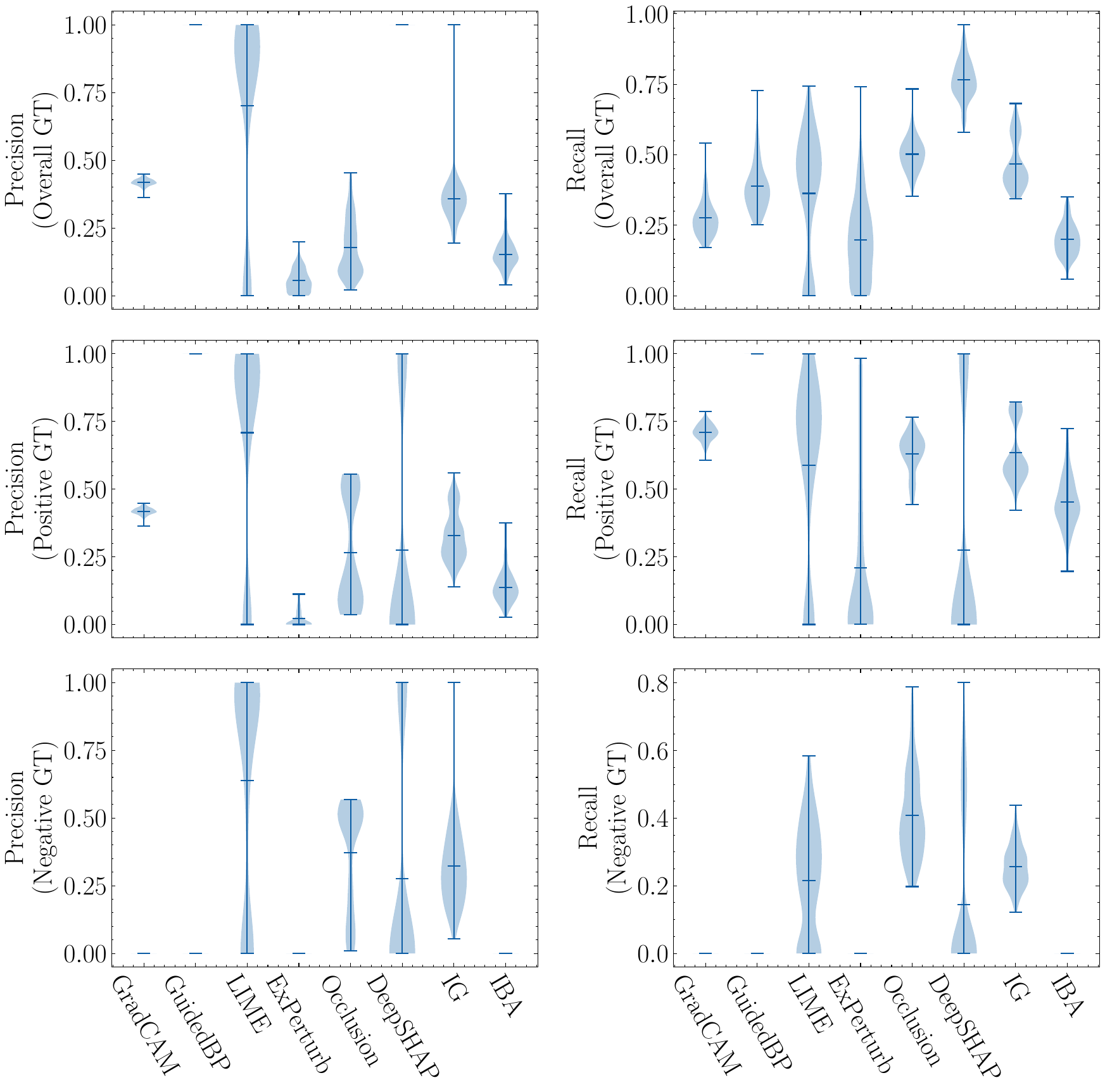}
    \caption{Ground-truth-based evaluation in the multi-color-sum setting. In each violin plot, the maximal, average, and minimal values are marked with horizontal bars. The first row. Note that only Occlusion, DeepSHAP, IG, and $\text{IG}^{*}$ yield negative attribution. For other attribution methods shown in the figure, we manually set their precision and recall associated with negative GT to $0.0$.}
    \label{fig:appendix:gt_eval_multi_color_sum_non_uniform}
\end{figure*}

Figure~\ref{fig:appendix:gt_eval_multi_color_sum_non_uniform} displays the violin plots of precisions and recalls computed with different sets of GT. Several methods, such as GradCAM, GuidedBP, LRP, and IBA, only identify the pixels contributing to the label class, ignoring the negatively contributing pixels, as shown in Figure~\ref{fig:appendix:visual_result_multi_color_sum}. Therefore, the recalls computed with positive GT for these methods are higher than the recalls computed with overall GT, as the overall GT includes both positively and negatively contributing pixels. Additionally, if we only consider the pixels of the label class, IBA exhibits superior performance by comparing the precision associated with positive GT in the multi-color-sum setting with the precision computed with overall GT in the single-color-modulo setting. A possible reason could be that IBA is formulated using variational methods, where the objective function is relaxed using cross-entropy. Cross-entropy is more suitable for multi-class classification in the multi-color-sum setting. To enhance performance in the single-color-modulo setting, an alternative version of the optimization objective for IBA might be required. Moreover, DeepSHAP, IG, and Occlusion necessitate knowing the ground truth baseline of the model, which is very difficult to acquire in real-world scenarios. In this study, we select a baseline $(0, 0, 0)$ that causes unpredicted model behavior, even though it is visually very close to the true baseline color $(20, 20, 20)$. In Figure~\ref{fig:appendix:visual_result_multi_color_sum}, we observe that Occlusion is prone to breaking down, and IG assigns random positive or negative attribution to the background. If we select the correct baseline for IG, we can significantly enhance its precision, as demonstrated by $\text{IG}^{*}$ in Figure~\ref{fig:appendix:gt_eval_multi_color_sum_non_uniform}. Furthermore, DeepSHAP achieves high precision and recall when computed with overall GT. However, it often misidentifies positively contributing pixels as negative, and vice versa. Consequently, DeepSHAP exhibits low precision and recall when computed with positive or negative GT. This suggests that DeepSHAP can accurately locate the ground truth features but often yields the wrong sign of attribution.

\begin{table*}[t]
    \centering
    \caption{Rankings of attribution methods in the multi-color-sum setting when evaluated with ground truth masks, Insertion/Deletion, and Sensitivity-N. Correlation denotes the Spearman's rank correlation with the \Fone-score (computed using the positive ground truth masks).}
    \begin{tabular}{c c c c c}
        \toprule
        \multirow{2}{*}{Attribution methods} & \multicolumn{4}{c}{Ranking} \\
        \cmidrule{2-5}
         & \Fone-score & Insertion & Deletion & Sensitivity-N \\
        \midrule
        GradCAM & 8 & 4 & 6 & 3 \\
        GuidedBP & 3 & 5 & 4 & 1 \\
        LIME & 1 & 3 & 3 & 2 \\
        ExPerturb & 4 & 6 & 7 & 8 \\
        Occlusion & 6 & 2 & 5 & 5 \\
        DeepSHAP & 5 & 8 & 2 & 6 \\
        IG & 7 & 1 & 1 & 4 \\
        IBA & 2 & 7 & 8 & 7 \\
        \midrule
        Correlation & $-$ & 0.02 & 0.47 & 0.65 \\
        \bottomrule
    \end{tabular}
    \label{tab:appendix:ranking_table_multi_color_sum}
\end{table*}

\subsection{Insertion/Deletion}
\label{sec:appendix:ins_del_multi_color_sum_non_uniform}
\subsubsection{Experimental setup}
The implementation of Insertion/Deletion in the multi-color-sum setting is consistent with the original paper. Specifically, we progressively perturb the pixels with replacement pixels with intensity $(0, 0, 0)$. This choice of replacement pixels aligns with standard practice. For Deletion, we compute the probability of the label class after perturbing the pixels in descending order sorted by attribution values. In the end, we obtain a curve of the predicted probability against the ratio of perturbed pixels. Then, we compute the area under the curve (AUC). A lower Deletion AUC indicates better performance of the attribution method. For Insertion, we compute the probability of the label class after inserting the pixels into a blank canvas with intensity $(0, 0, 0)$ in ascending order sorted by attribution values. Again, the predicted probability after insertion is recorded, allowing us to obtain a curve of the predicted probability against the ratio of inserted pixels. Then, we compute the AUC, and a higher Insertion AUC indicates better performance of the attribution method.

\subsubsection{Experimental results} 

As shown in Table~\ref{tab:appendix:ranking_table_multi_color_sum} and Figure~\ref{fig:main:ins_curves_multi_color_sum_non_uniform}, Insertion/Deletion display limited consistency with GT-based evaluation, when the Unseen Data Effect is present.


\subsection{Sensitivity-N}
\label{sec:appendix:sens_n_multi_color_sum_non_uniform}
\subsubsection{Experimental setup}
The implementation of Sensitivity-N in the multi-color-sum setting is consistent with the original paper. For a specific number of perturbed pixels $N$, we randomly select $N$ pixels in the image, and perturb them to the default baseline $(0, 0, 0)$. After that, we feed the perturbed image into the neural network and obtain the predicted probability of the label class. Additionally, we record the sum of attribution of the perturbed pixels. This process is repeated for 100 times, yielding a pair of vectors with 100 elements, where the first vector represents the predicted probability in 100 repetitions and the second vector represents the sum of attribution. Next, we compute the Pearson Correlation Coefficient between the vectors of predicted probability and sum of attribution. With $N$ increasing from 1 to $224 \times 224$, we obtain a curve of Pearson Correlation Coefficient. The greater the correlation is, the better the attribution method performs.

\subsubsection{Experimental results} 
\begin{figure}[t]
    \centering
    \includegraphics[width=\columnwidth]{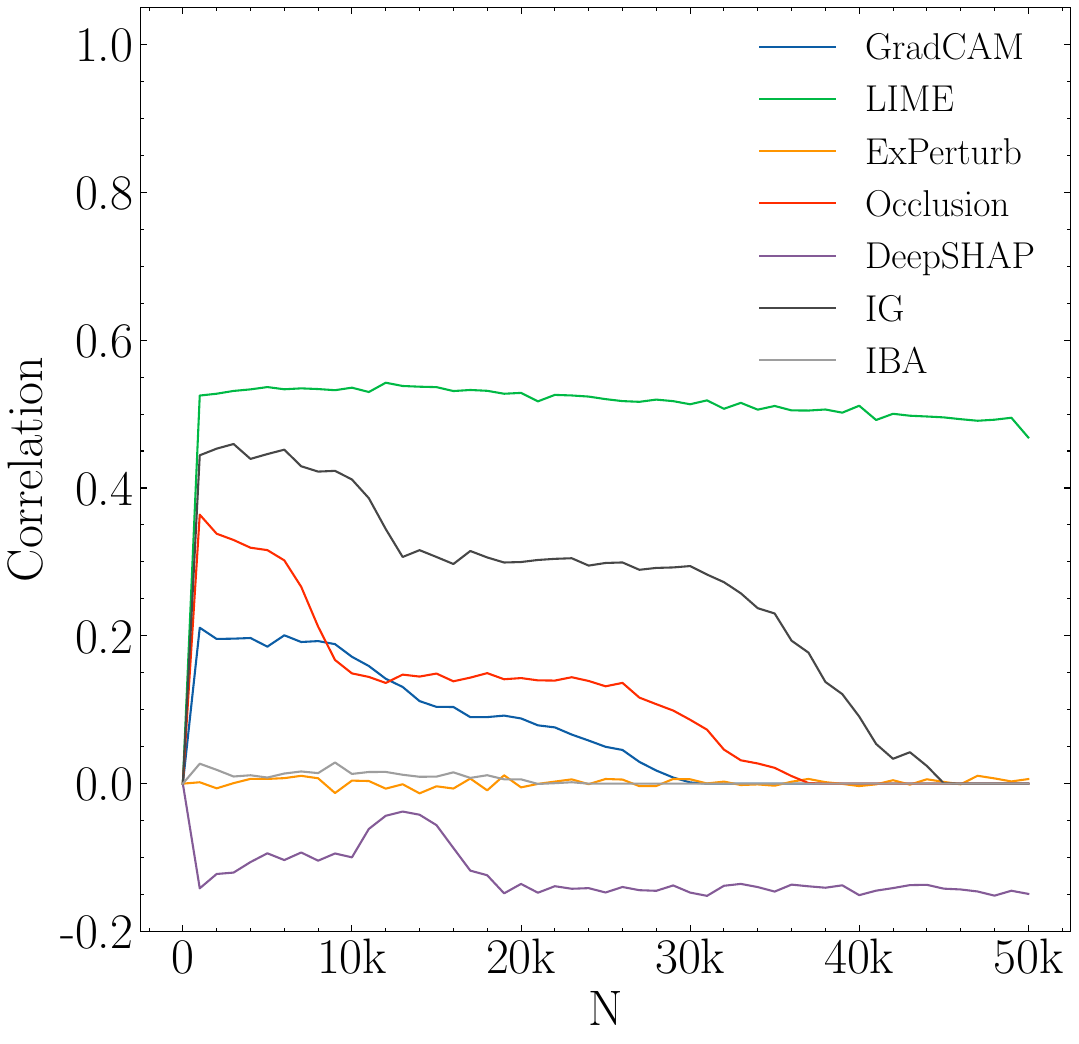}
    \caption{Sensitivity-N in the multi-color-sum setting.}
    \label{fig:main:sens_n_multi_color_sum_non_uniform}
\end{figure}
The Sensitivity-N curves are shown in Figure~\ref{fig:main:sens_n_multi_color_sum_non_uniform}.
First, we observe that the performance of the attribution methods is much less distinguishable than in the single-color-modulo case. For instance, the curves of GradCAM, GuidedBP, and Occlusion intersect with one another multiple times, making it harder to conclude which method displays superior performance. Second, DeepSHAP produces a negative correlation curve. This can be interpreted by our previous analysis that DeepSHAP often misidentifies the sign of contributing pixels. Therefore, the sum of attribution of perturbed pixels is much less aligned with the predicted probability of the perturbed image. Third, in order to rank these methods and compare the ranking with the ranking on GT-based evaluation, we simply compute the average correlation over all $N$s. As demonstrated in Table~\ref{tab:appendix:ranking_table_multi_color_sum}, the Sensitivity-N evaluation result is much less correlated with the ground-truth-based evaluation compared to that in the single-color-modulo setting.

\section{Experiment Setting for Figure~\ref{fig:main:visualize_multi_color_sum}}\label{sec:appendix:setting_for_visualization}
All attribution maps generated in the same setting that is discussed in Appendix~\ref{sec:appendix:exp_multi_color_sum_non_uniform}. 

\section{Experiments in the multi-color-sum setting (without Unseen Data Effect)} \label{sec:appendix:exp_multi_color_sum_uniform}

In our default settings for the Accumulator module within the multi-color setting, we employ a non-uniform weight initialization. In this section, we present supplementary empirical results from the multi-color-sum setting, where we implement a uniform weight initialization in the Accumulator. Please note that: (1) the weight initialization schemes do not impact the Accumulator output, ensuring that the model still achieves $100\%$ accuracy on the synthetic dataset; and (2) the color detector retains redundant channels and consequently, remains susceptible to Unseen Data Effect. In the following subsections, the evaluation settings are identical as the Appendix~\ref{sec:appendix:exp_multi_color_sum_non_uniform}, except for the number of redundant channels in the Accumulator. 

\begin{table*}[h]
    \centering
    \caption{Rankings of attribution methods in the multi-color-sum setting (without Unseen Data Effect). Attribution methods are evaluated with ground truth masks, Insertion/Deletion, and Sensitivity-N. Correlation denotes the Spearman's rank correlation with the \Fone-score (computed using the overall ground truth masks).}
    \begin{tabular}{c c c c c}
        \toprule
        \multirow{2}{*}{Attribution methods} & \multicolumn{4}{c}{Ranking} \\
        \cmidrule{2-5}
         & \Fone-score & Insertion & Deletion & Sensitivity-N \\
        \midrule
        GradCAM & 5 & 5 & 6 & 6 \\
        GuidedBP & 6 & 6 & 4 & 1 \\
        LIME & 1 & 4 & 3 & 4 \\
        ExPerturb & 7 & 8 & 7 & 8 \\
        Occlusion & 2 & 3 & 5 & 5 \\
        DeepSHAP & 3 & 1 & 2 & 2 \\
        IG & 4 & 2 & 1 & 3 \\
        IBA & 8 & 7 & 8 & 7 \\
        \midrule
        Correlation & $-$ & 0.42 & 0.61 & 0.81 \\
        \bottomrule
    \end{tabular}
    \label{tab:appendix:ranking_table_w_ig_star_multi_color_sum_wo_ood}
\end{table*}
Based on the data provided in Table~\ref{tab:appendix:ranking_table_w_ig_star_multi_color_sum_wo_ood}, it's apparent that the correlations between ground-truth-based evaluation and other metrics do not show significant improvement. This suggests that simplifying the Accumulator network by utilizing uniform weights does not necessarily mitigate the adverse behavior of Insertion/Deletion and Sensitivity metrics that stem from Unseen Data Effect.

\begin{figure}[h!]
    \centering
    \includegraphics[width=0.9\columnwidth]{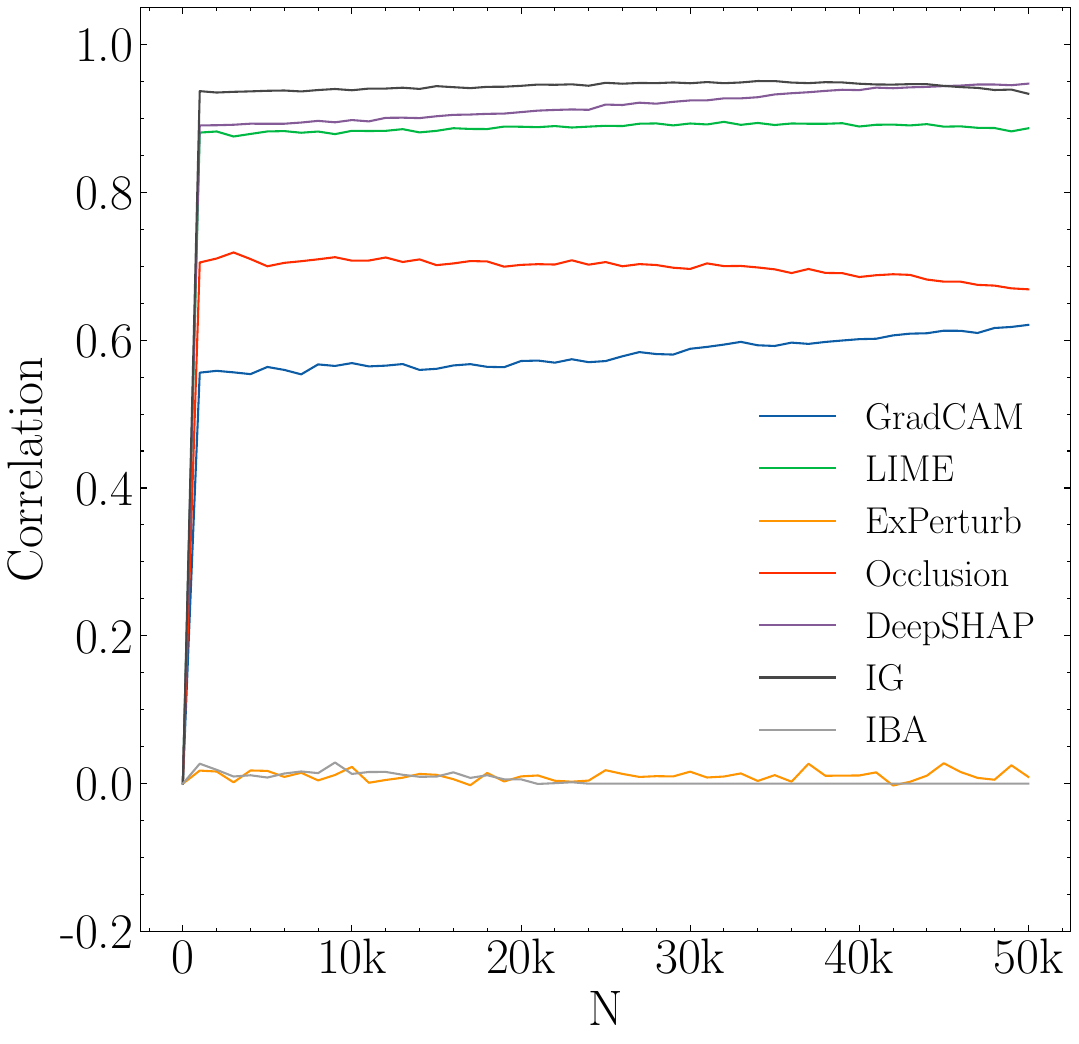}
    \caption{Sensitivity-N in multi-color-sum setting without the Unseen Data Effect}
    \label{fig:main:sens_n_multi_color_sum_wo_ood}
\end{figure}

\subsection{Ground-truth-based evaluation}
\begin{figure*}[t]
    \centering
    \includegraphics[width=\textwidth]{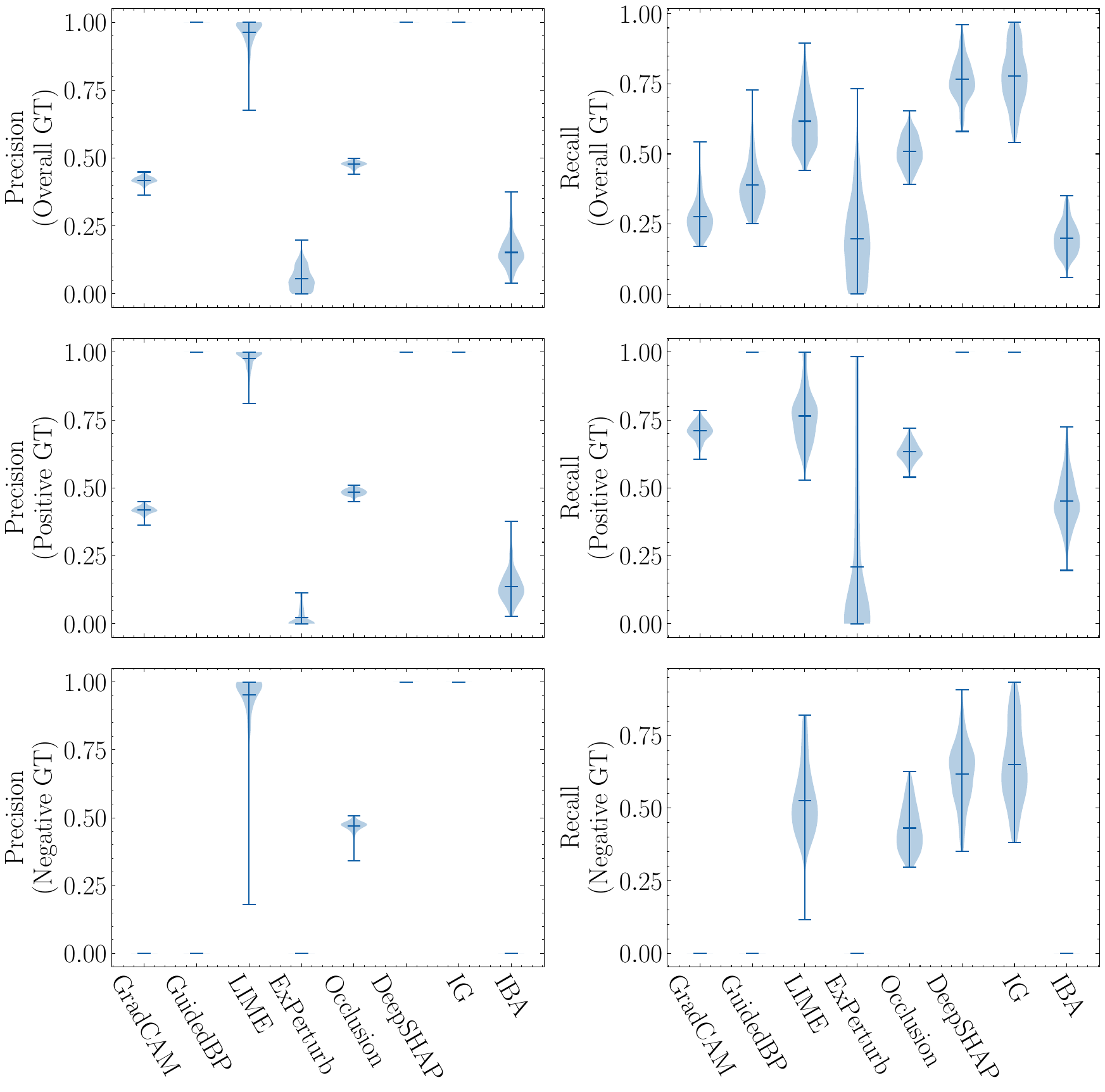}
    \caption{Ground-truth-based evaluation results in the multi-color-sum setting without Unseen Data Effect.}
    \label{fig:appendix:gt_eval_multi_color_sum_wo_ood}
\end{figure*}

Figure~\ref{fig:appendix:gt_eval_multi_color_sum_wo_ood} illustrates the violin plots of various attribution methods, evaluated with different sets of GT.
\subsection{Insertion/Deletion}
Figure~\ref{fig:main:ins_curves_multi_color_sum_wo_ood} shows the Insertion curve.

\subsection{Sensitivity-N}
Figure~\ref{fig:main:sens_n_multi_color_sum_wo_ood} depicts the correlation curves.

\section{Broader impacts}
In this work, we introduce an evaluation framework designed specifically for feature attribution methods. A notable gap in the research community has been the absence of reliable evaluation metrics, leading to various complications. For example, numerous feature attribution methods have been proposed that, in some cases, even yield inconsistent attribution results given identical inputs. Our approach guarantees its validity for evaluation purposes because of the fully synthetic nature. Our synthetic settings may appear primitive compared to real-world datasets and neural networks trained on such datasets. However, they provide a controlled laboratory environment, enabling a thorough examination of feature attribution methods prior to deployment. This facilitates a robust evaluation and refinement process for attribution methods.

\section{Additional study on segmentation methods of LIME}

\begin{figure*}[t]
    \centering
    \includegraphics[width=0.6\textwidth]{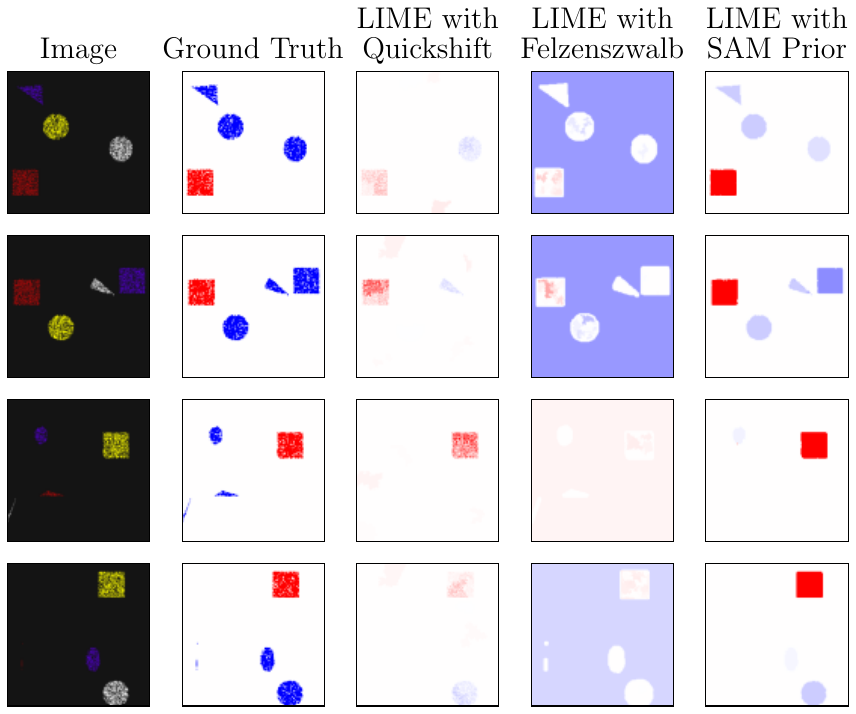}
    \caption{LIME in the multi-color-sum setting with different segmentation methods. In contrast to Quickshift \citep{vedaldi2008quick} and Felzenszwalb \citep{felzenszwalb2004efficient}, SAM \citep{kirillov2023sam} offers markedly improved segmentation masks for inputs. This improvement significantly boosts the faithfulness of attribution maps, ensuring closer alignment with the ground truth attribution map.}
    \label{fig:appendix:lime_sam_multi_color_sum}
\end{figure*}

\begin{figure*}[t]
    \centering
    \includegraphics[width=0.6\textwidth]{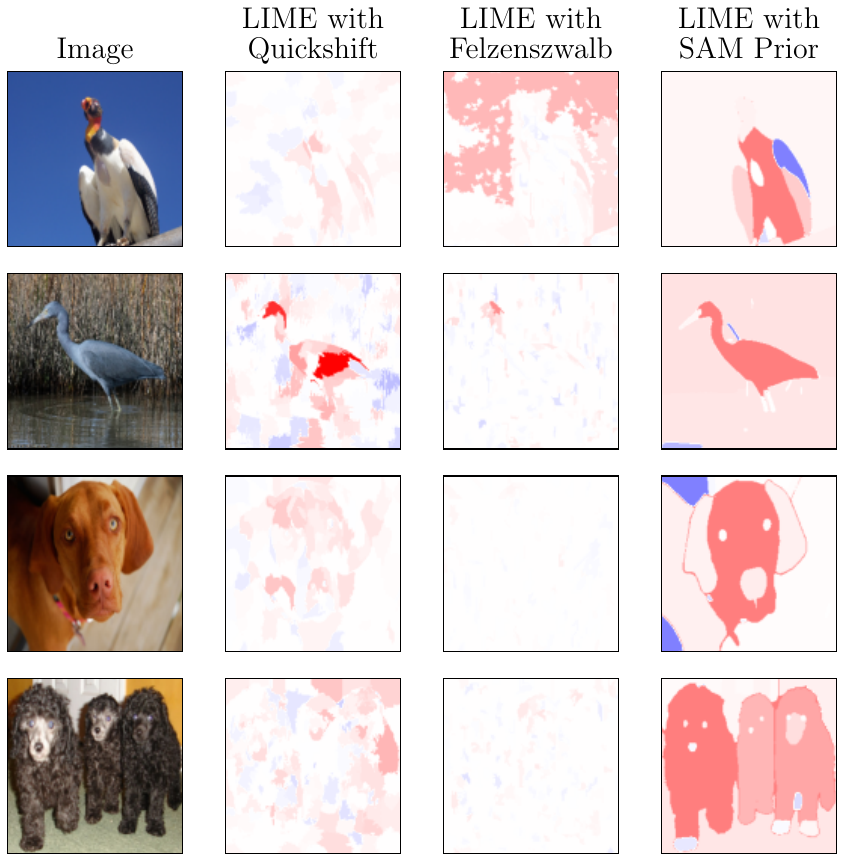}
    \caption{LIME attribution maps for VGG16 with different segmentation methods. In line with results in the multi-color-sum setting, SAM segments input features into more compact and semantically meaningful groups. Consequently, the LIME attribution maps generated using SAM exhibit reduced noise and are visually more congruent with human perception.}
    \label{fig:appendix:lime_sam_imagenet}
\end{figure*}

\begin{figure*}[t]
    \centering
    \includegraphics[width=0.6\textwidth]{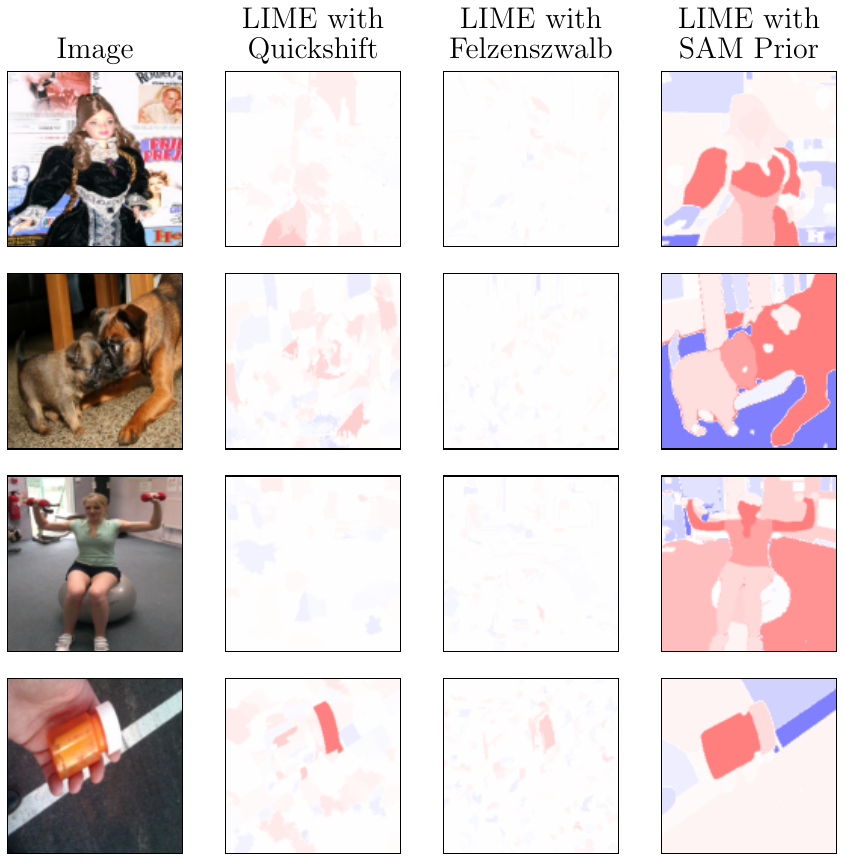}
    \caption{LIME attribution maps for ViT-B-16 with different segmentation methods. These attribution maps show a significant enhancement in attribution quality when LIME incorporates SAM, and this finding generalizes to vision transformers.}
    \label{fig:appendix:lime_sam_vit_imagenet}
\end{figure*}

In Section~\ref{sec:exp:faithfulness_lime}, we observed that the segmentation method significantly affects LIME's performance. This observation prompted us to explore a more advanced segmentation approach capable of effectively grouping features that function together and identifying independent feature groups. We conducted a comparative analysis of LIME attribution maps generated using Quickshift~\citep{vedaldi2008quick}, Felzenszwalb~\citep{felzenszwalb2004efficient}, and the recent segmentation approach Segment Anything Model (SAM)~\citep{kirillov2023sam}. Our comparison in the multi-color-sum setting, as shown in Figure~\ref{fig:appendix:lime_sam_multi_color_sum}, reveals a substantial increase in faithfulness when SAM is utilized as the segmentation method. 
Furthermore, we employ LIME for explaining two ImageNet-pretrained models: VGG~\citep{simonyan2015a} and ViT~\cite{dosovitskiy2020image}, respectively.
Visual inspection of the attribution maps on VGG16 (Figure~\ref{fig:appendix:lime_sam_imagenet}) and ViT-B-16 (Figure~\ref{fig:appendix:lime_sam_vit_imagenet}), indicates that SAM provides a prior of more cohesively grouped features and significantly reduces noise in the LIME attribution maps. These attribution maps are much better aligned with human perception as well. Moreover, this finding is consistent across the CNN and ViT architectures. This study underscores that the findings in AttributionLab can assist researchers to identify strategies for enhancing existing attribution methods, and it demonstrates that these enhancements can generalize to complex real-world scenarios.

\end{document}